\newif\ifconfver
\confverfalse      
\confvertrue        

\newif\ifonecoltab
\onecoltabtrue        

\newif\ifplainver  
\plainvertrue

\ifplainver
    \confverfalse                         
\fi

\ifconfver
     \documentclass[10pt,twocolumn,twoside]{IEEEtran}
\else
    \ifplainver
        \documentclass[11pt]{article}
        \usepackage{fullpage}
    \else
        \documentclass[12pt,draftcls,onecolumn]{IEEEtran}
    \fi
\fi

\usepackage{calc,amsfonts,amssymb,amsmath,bm,url,color,theorem,graphicx,cite}
\usepackage{psfrag,subfigure,float}
\usepackage[algoruled,linesnumbered]{algorithm2e}
\usepackage{multirow}

\definecolor{orange}{RGB}{255,107,0}

\newtheorem{Lemma}{Lemma}
\newtheorem{Prop}{Proposition}

\theorembodyfont{\rmfamily}

\newtheorem{Remark}{Remark}

\begin{document}

\bibliographystyle{IEEEtran}

\newcommand{\papertitle}{
Semiblind Hyperspectral Unmixing in the Presence of Spectral Library Mismatches
}

\newcommand{\paperabstract}{
The dictionary-aided sparse regression (SR) approach has recently emerged as a promising alternative to hyperspectral unmixing (HU) in remote sensing.
By using an available spectral library as a dictionary,
the SR approach identifies the underlying materials in a given hyperspectral image by selecting a small subset of spectral samples in the dictionary to represent the whole image.
A drawback with the current SR developments is that an actual spectral signature in the scene is often assumed to have zero mismatch with its corresponding dictionary sample,
and such an assumption is considered too ideal in practice.
In this paper, we tackle the spectral signature mismatch problem by proposing a dictionary-adjusted nonconvex sparsity-encouraging regression (DANSER) framework.
The main idea is to incorporate dictionary correcting variables in an SR formulation.
A simple and low per-iteration complexity algorithm is tailor-designed for practical realization of DANSER.
Using the same dictionary correcting idea, we also propose a robust subspace solution for dictionary pruning.
Extensive simulations and real-data experiments show that the proposed method is effective in mitigating the undesirable spectral signature mismatch effects.
}


\ifplainver

    \date{June 24, 2015}

    \title{\papertitle \footnote{Part of this work was published in WHISPERS 2014 \cite{fu2013greedy}.}}

    \author{
    $^\ast$Xiao Fu, $^\dag$Wing-Kin Ma, $^\star$Jos\'{e} M. Bioucas-Dias, and $^\ddag$Tsung-Han Chan
    \\ ~ \\
		$^\ast$Department of Electrical and Computer Engineering, University of Minnesota,\\
		Minneapolis, 55455, MN, United States\\
		Email: xfu@umn.edu
		\\~\\
    $^\dag$Department of Electronic Engineering, The Chinese University of Hong  Kong, \\
    Hong Kong \\
    Email: wkma@ieee.org
    \\ ~ \\
    $^\star$ Instituto de Telecomunica\c{c}\~{o}es and
    Instituto Superior T\'{e}cnico, \\
    1049-1, Lisbon, Portugal\\
		Email: bioucas@lx.it.pt
		\\ ~ \\
    $^\ddag$ MediaTek Inc., Hsinchu, Taiwan \\
    Email: thchan@ieee.org
		\\
    }

    \maketitle

    \begin{abstract}
    \paperabstract
    \end{abstract}

\else
    \title{\papertitle}

    \ifconfver \else {\linespread{1.1} \rm \fi

   \author{Xiao Fu, Wing-Kin Ma, Jos\'{e} M. Bioucas-Dias, and Tsung-Han Chan
\thanks{Part of this work was published in WHISPERS 2014 \cite{fu2014robust}.
X. Fu is with the Department of Electrical and Computer Engineering, University of Minnesota, Minneapolis, MN, 55455, e-mail xfu@umn.edu.
W.-K. Ma is with the Department of Electronic Engineering, the Chinese University of Hong Kong, e-mail wkma@ee.cuhk.edu.hk.
J.M. Bioucas-Dias is with the Instituto de Telecomunica\c{c}\~{o}es and Instituto Superior T\'{e}cnico, 1049-1, Lisbon, Portugal, e-mail bioucas@lx.it.pt.
T.-H. Chan is with the MediaTek Inc., Hsinchu, Taiwan, e-mail thchan@ieee.org.}
}

    \maketitle

    \ifconfver \else
        \begin{center} \vspace*{-2\baselineskip}
        \end{center}
    \fi

    \begin{abstract}
    \paperabstract
    \end{abstract}

    \begin{keywords}\vspace{-0.0cm}
        Semibind hyperspectral unmixing, compressive sensing, dictionary mismatch, $\ell_p$ quasi-norm sparsity promoting, robust dictionary pruning
    \end{keywords}

    \ifconfver \else \IEEEpeerreviewmaketitle} \fi

 \fi

\ifconfver \else
    \ifplainver \else
        \newpage
\fi \fi

\section{Introduction}
Hyperspectral unmixing (HU) aims at decomposing pixels of an hyperspectral image (HSI) into constituent spectra that represent some pure materials.
HU is useful in a number of applications, such as environment surveillance, agriculture, mine detection, and food and medicine analytics.
As one of the core developments in signal and image processing for HSIs, various HU algorithms have been developed in the past two decades from different perspectives,
such as Bayesian inference, nonnegative matrix factorization, convex analysis, pure pixel pursuit, and many more; see, e.g., \cite{Bioucas2012,Ma2014} for some overviews.

Recently,
a class of HU algorithms based on spectral libraries
has attracted much attention.
A spectral library is a collection of spectral signatures of materials acquired in controlled or ideal environments, e.g., in laboratories.
There are several publicly available libraries, provided by government agencies and research institutes.
For example, the U.S. Geological Survey (U.S.G.S.) library \cite{USGS2007} contains remotely sensed and extracted spectral signatures of over $1300$ materials.
Such rich knowledge of materials' spectra in the existing libraries provides new opportunities for HU.
By using an existing library as a dictionary, and by assuming the linear mixture model,
we can treat HU as a problem of selecting a small number of spectra from the dictionary to represent all the  pixels.
Such a dictionary-aided semiblind formulation is fundamentally identical to the well-known basis selection or sparse regression problem in compressive sensing (CS),
and thus many well-developed tools from CS can be applied.
Fundamentally, there are several advantages with dictionary-aided semiblind HU.
First, unlike many blind HU approaches (which do not use dictionaries),
dictionary-aided methods do not require assumptions
such as the pure pixel assumption and the sum-to-one abundance conditions.
Second, dictionary-aided methods may not require knowledge of the number of materials contained in the HSIs.

Several dictionary-aided HU algorithms based on sparse regression were proposed in \cite{Iordache2011,Iordache2012TV,Iordache2012collaborative,joseMUSIC2013}.
The algorithms in \cite{Iordache2011,Iordache2012TV} and \cite{Iordache2012collaborative} treat the HU problem as a single pixel-based sparse regression (SR) problem and
a multiple pixel-based collaborative sparse regression (CSR) problem, respectively.
Classic $\ell_1$ norm and $\ell_2/\ell_1$ mixed-norm minimization-based sparse optimization methods are employed to tackle the formulated problems there.
The corresponding optimization problems are convex, and thus can be solved efficiently, e.g., by some specialized alternating direction method of multipliers (ADMM)-based algorithms.
Three main difficulties have been observed when applying the algorithms in \cite{Iordache2011,Iordache2012TV,Iordache2012collaborative}, however:
First, the spectral library members (i.e., the recorded material spectra) exhibit very high mutual coherence.
As is known in CS \cite{Tropp2006pt2,Eldar2010average,Chen2006,tropp2006just},
high mutual coherence may lead to poor performance when applying $\ell_1$ norm and $\ell_2/\ell_1$ mixed-norm minimization-based sparse optimization.
Second, the size of a spectral library is often very large.
Consequently, we are faced with a large-scale problem, for which computational efficiency becomes an issue.
Third, there may be mismatches between the actual spectral signatures in the scene and the dictionary samples, due to various reasons.
Such dictionary mismatches affect the performance of a dictionary-aided semiblind HU to an extent which depends on the severity of the mismatches.
%

The first two difficulties mentioned above have been
tackled
by employing a dictionary pruning method based on multiple signal classification (MUSIC) \cite{joseMUSIC2013}.
MUSIC is a classical subspace method in sensor array processing \cite{schmidt1986multiple}, and recently finds its application in CS \cite{kim2010compressive}.
In dictionary-aided semiblind HU, MUSIC proves to be useful in pre-selecting some relevant spectra from a large spectral library.
As a result, a size-reduced dictionary can be constructed for the SR and CSR algorithms to perform semiblind HU.
After dictionary pruning, both the mutual coherence of the dictionary  and the complexity of the subsequent semiblind HU algorithm can be reduced.

However, the third difficulty, spectral signature mismatches, is still not addressed.
In practice, the mismatch problem arises for several reasons.
First, the materials' spectra may vary from time to time, and from site to site,
subject to diverse physical conditions, e.g. strength of sunlight and temperature \cite{somers2011endmember}.
Second, the calibration procedure for spectral signatures may introduce errors.
Third, the spatial resolutions of spectra in the dictionary can be different from those of the image, and that can also result in modeling errors.
Spectral mismatches can be rather damaging to the existing semiblind HU algorithms;
particularly, MUSIC-based dictionary pruning is sensitive to spectral signature mismatches, as will be seen in the simulations.

\bigskip

\noindent
\textbf{Contributions}
In this work, we propose a dictionary-aided HU framework that takes spectral signature mismatches into consideration.
Our first contribution lies in developing a new dictionary-aided HU algorithm.
The formulation leading to the new algorithm uses insights of CSR,
but has two key differences: 1) We model spectral signature mismatches as bounded error vectors, and
attempt to compensate those errors in the formulation.
2) We employ the nonconvex $\ell_2/\ell_p$ ($0<p<1$) quasi-norm as the sparsity-promoting function, instead of the convex $\ell_2/\ell_1$ mixed norm as in CSR \cite{Iordache2012collaborative}.
The second endeavor is motivated by the fact that
quasi-norm based
sparse optimization
has been demonstrated to exhibit better sparsity promoting performance in certain difficult situations, e.g., the high-coherence dictionary case \cite{Chartrand2008,Rao1999,chartrand2008restricted}.
Since our formulation considers dictionary adjustment, it is more complicated to handle than the previous CSR work.
We derive the new algorithm by careful design of alternating optimization,
and its upshot is that the solution update at each iteration involves simple matrix operations.


The second contribution is a spectral mismatch-robust solution to dictionary pruning.
We give a robust MUSIC formulation, wherein the goal is to identify spectral signature samples that are close to the true materials' signatures, rather than being exactly equal.
At first look, the robust MUSIC method seems to be computationally expensive under large dictionary sizes; specifically, for every dictionary sample, we need to solve an optimization problem.
We show that, however, the optimization problem in robust MUSIC can be converted to a single-variable optimization problem, thereby being solved with a very low computational cost.
Simulations and real data experiment are used to show the effectiveness of the proposed algorithm.

~\\
\textbf{Related Works}:
While the topic of CS and sparse regression has received enormous attention in various fields, there are comparatively fewer works that study sparse regression in the presence of dictionary mismatches. Those works usually appear in signal processing, and the application is not HU.
In \cite{zhu2011sparsity}, perturbations of dictionaries were modeled as Gaussian noise, and an $\ell_1$-norm regularized total least squares criterion was proposed; there, the focus was the single-measurement vector case (or the single-pixel case in our problem), and constraints on the unknowns were not considered.
In \cite{gribonval2012blind,bilen2013blind}, dictionary perturbations were modeled as scaling factors on each dictionary atom,
and the formulated problem is convex.
The algorithm in \cite{tan2014joint} attacked the dictionary mismatch problem in CSR-based direction-of-arrival finding.
There, the mismatch was characterized by a subspace of a structured matrix, and the optimization surrogates there are also convex $\ell_2/\ell_1$ norm and its smoothed counterparts.
We also note that $\ell_p$ quasi-norm based sparse regression was applied to HU for single pixel-based unmixing without considering dictionary mismatches \cite{chen2013sparse}.
Here, our focus is collaborative sparse regression using multiple pixels, which is known to have both theoretical and practical advantages over the single pixel-based algorithms;
we adopt the nonconvex $\ell_2/\ell_p$ quasi-norm, where $0<p<1$, as our sparsity-promoting function, since it has proven to show better performance in various applications;
and we model spectral mismatches as deterministic bounded errors,
which does not require statistical assumptions and may be more flexible.

~\\
\emph{Notation}:
The notations ${\bm x} \in \mathbb{R}^n$ and ${\bm X} \in \mathbb{R}^{m \times n}$ mean that ${\bm x}$ and ${\bm X}$ are a real-valued $n$-dimensional vector and a real-valued $m \times n$ matrix, respectively (resp.).
The notation ${\bm x} \geq {\bm 0}$ (resp. ${\bm X} \geq {\bm 0}$) means that ${\bm x}$ (resp. ${\bm X}$) is element-wise non-negative.
The $i$th column of a matrix ${\bm X} \in \mathbb{R}^{m \times n}$ is denoted by ${\bm x}_i \in \mathbb{R}^m$,
and the $j$th row of ${\bm X}$ is denoted by ${\bm x}^j$.
The superscript ``$T$'' and ``$-1$'' stand for the transpose and inverse operations, resp.
The orthogonal projector of the range space of $\bm X$ is denoted by ${\bm P}_{\bm X} = {\bm X} ( {\bm X}^T {\bm X})^\dag {\bm X}^T$,
where the superscript ``$\dag$'' stands for the pseudo-inverse;
and the corresponding orthogonal complement projector is denoted by ${\bm P}_{\bm X}^\perp = {\bm I} - {\bm P}_{\bm X}$.
The $\ell_p$ norm of a vector ${\bm x}\in\mathbb{R}^n$, $p \geq 1$, is denoted by $\| {\bm x}\|_p = (\sum_{i=1}^n |x_i|^p)^{1/p}$.
The $\ell_p$ quasi-norm, $0 < p < 1$, is denoted by the same notation as above.
The mixed $\ell_p/\ell_q$-norm or $\ell_p/\ell_q$-quasi-norm is denoted by $\|{\bm X}\|_{q,p}= ( \sum_{i=1}^m\|{\bm x}^i\|_q^p )^{1/p}$.
The Frobenious norm is denoted by $\|{\bm X}\|_F = \| {\bm X} \|_{2,2}$.

\section{Background}

\subsection{Signal Model and Dictionary-Aided Semiblind HU}

Consider a remotely sensed scene that is composed of mixtures of $N$ different materials.
Assuming linear mixtures,
the measured hyperspectral image can be modeled as
\begin{equation}
\begin{aligned}
          {\bm y}[\ell] = \sum_{n=1}^N {\bm a}_n{s}_n[\ell] + {\bm v}[\ell],\quad \ell= 1,\ldots,L,
\end{aligned}	\label{eq:LMM}
\end{equation}
where ${\bm y}[\ell]\in\mathbb{R}^M$ denotes the hyperspectral measurement at the $\ell$th pixel of the image,
with $M$ being the number of spectral bands;
each ${\bm a}_n\in{\mathbb{R}^M}$, $n=1,\ldots,N$, represents the spectral signature of a particular material, indexed by $n$ here;
$s_n[\ell] \geq 0$ is the abundance of material $n$ at pixel $\ell$;
${\bm v}[\ell]\in\mathbb{R}^M$ is a noise vector;
and $L$ is the number of pixels.
For convenience, we will write \eqref{eq:LMM} in a matrix form
\begin{equation}
{\bm Y} = {\bm A}{\bm S} + {\bm V},
\label{eq:LMM_blk}
\end{equation}
where
${\bm Y}=[{\bm y}[1],\ldots,{\bm y}[L]]$,
${\bm A}=[{\bm a}_1,\ldots,{\bm a}_N]$,
${\bm S}=[{\bm s}[1],\ldots,{\bm s}[L]]$,
${\bm s}[\ell]=[s_1[\ell],\ldots,{s}_N[\ell]]^T$,
and ${\bm V}=[{\bm v}[1],\ldots,{\bm v}[L]]$.

In HU, we aim by identifying ${\bm A}$ and ${\bm S}$ from ${\bm Y}$.
This amounts to a blind separation problem where hyperspectral signal-specific properties---such as pure pixel and sum-to-one abundance conditions---are often utilized to attack the problem in many existing and concurrent HU studies.
Dictionary-aided semiblind HU takes a different strategy.
Motivated by the fact that many spectral libraries (e.g., the U.S.G.S. library \cite{USGS2007}) have been built up in the past decades,
its principle is to use one such spectral library as a dictionary to infer what are the underlying spectral signatures, and hence materials, in the scene.
To put this into context,
define
\[{\bm D} = [{\bm d}_1,\ldots,{\bm d}_K]\in\mathbb{R}^{M\times K}\]
as a spectral dictionary,
where each ${\bm d}_k\in\mathbb{R}^M$ is a previously recorded spectral sample for a specific material, and $K$ denotes the dictionary size or the number of spectral samples.
A dictionary often contains a wide variety of samples of materials, and as such $K$ is large.
The key assumption with dictionary-aided semiblind HU is that the dictionary covers the spectral signatures of all materials in the scene; that is to say,
\[{\bm a}_n\in\{{\bm d}_1,\ldots,{\bm d}_K\}, \quad \text{for every $n=1,\ldots,N$.} \]
Alternatively, we can write, for each $n=1,\ldots,N$,
\begin{equation} \label{eq:exact_match}
{\bm a}_n = {\bm d}_{k_n}, \quad \text{for some $k_n \in \{ 1,\ldots, K \}$.}
\end{equation}
Consequently, the signal model in \eqref{eq:LMM_blk} can be written as
\begin{equation}
{\bm Y} = {\bm D}{\bm C} + {\bm V},
\label{eq:LMM_blk_sparse}
\end{equation}
where ${\bm C} \in \mathbb{R}^{K,L}$ is a row-sparse matrix;
to be specific, the $k_n$th row of ${\bm C}$, $n=1,\ldots,N$, is the $k$th row of ${\bm S}$, and the other rows of ${\bm C}$ are all zeros.

Let us consider the sparse regression approach---currently the main approach for dictionary-aided semiblind HU.
The idea is to exploit the sparsity of ${\bm C}$, thereby attempting to recover the indices $k_1,\ldots,k_n$ correctly and the abundance matrix ${\bm S}$ accurately.
There is more than one way to formulate such a sparse promoting problem (see, e.g., \cite{Bioucas2012,Ma2014} and the references therein), and here we are interested in the CSR formulation~\cite{Iordache2012collaborative,joseMUSIC2013}.
The CSR formulation is given as follows:
\begin{equation}
\begin{aligned}
\min_{{\bm C}\in\mathbb{R}^{K\times L}}~&\|{\bm Y}-{\bm D}{\bm C}\|_F^2 + \lambda \|{\bm C}\|_{2,1}\\
{\rm s.t.}~&{\bm C}\geq{\bm 0},
\end{aligned}\label{eq:CSR}
\end{equation}
for some prespecified constant $\lambda > 0$.
Here, notice that $\|{\bm C}\|_{2,1} = \sum_{i=1}^K \| {\bm c}^i \|_2$, which aims at promoting row sparsity of ${\bm C}$.
As can be seen in Problem~\eqref{eq:CSR},
CSR seeks to find a nonnegative row-sparse ${\bm C}$ that provides a good approximation to ${\bm Y} = {\bm D}{\bm C}$.
Problem~\eqref{eq:CSR} is convex,
and a fast algorithm based on ADMM has been derived for Problem~\eqref{eq:CSR}~\cite{Iordache2012collaborative}.

\subsection{Dictionary Pruning using the Subspace Approach}

As discussed in the Introduction,
large dictionary size and high mutual coherence with the dictionary are two main difficulties encountered in CSR and other sparse regression methods,
and these two difficulties may be circumvented by applying dictionary pruning.
Here, we are interested in a subspace-based dictionary pruning method called MUSIC~\cite{joseMUSIC2013}.
This subspace method may be best described by studying the noiseless case ${\bm Y} = {\bm A}{\bm S}$.
Let ${\bm U}_S\in\mathbb{R}^{M\times N}$ denote a matrix that contains the first $N$ left singular vectors of ${\bm Y}$.
It can be shown that in the noiseless case and under some mild assumptions\footnote{Specifically, we require that ${\bm S}$ has full row rank, and that ${\rm spark}({\bm D}) > N+1$, where ${\rm spark}({\bm X})=r$ means that any $r$ columns of ${\bm X}$ are linearly independent. Intuitively, these requirements mean that the abundance maps of the different materials are sufficiently different, and that any $N$ spectral samples in the dictionary are sufficiently different.}, we have
\begin{equation}
        {\bm P}^{\perp}_{{\bm U}_S}{\bm d}_k = {\bm 0} \Longleftrightarrow
\text{ ${\bm d}_k = {\bm a}_{k_n}$ for some $n \in \{1,\ldots,N \}$.}
\label{eq:suff_necc}
\end{equation}
The physical meaning of \eqref{eq:suff_necc} is that if a spectral sample ${\bm d}_k$ in the dictionary is also one of the spectral signatures in the scene, then it must be perpendicular to the orthogonal complement signal subspace.
Also, the converse is true.
From an algorithm viewpoint,
the above observation suggests that we can correctly identify the indices $k_1,\ldots,k_N$ by the simple closed-form equations at the left-hand side (LHS) of \eqref{eq:suff_necc}---at least in the noiseless case.

In practice, where noise is present, the LHS of \eqref{eq:suff_necc} may not be exactly all-zero.
Under such circumstances, the following procedure can be used to estimate $k_1,\ldots,k_N$:
\begin{enumerate}
  \item For $k=1,\ldots,K$, calculate
                \begin{align} \label{eq:MUSIC}
                             \gamma_{\rm MUSIC}(k) = \frac{{\bm d}_k^T{\bm P}^{\perp}_{{\bm U}_S}{\bm d}_k}{\|{\bm d}_k\|_2^2}.
                \end{align}
   \item Determine $\hat{\Lambda}=\{  \hat{k}_1,\ldots, \hat{k}_N  \}$ such that for $n=1,\ldots,N$, we have $\gamma_{\rm MUSIC}( \hat{k}_n)<\gamma_{\rm MUSIC}(j)$ for all $j\notin\hat{\Lambda}$.
\end{enumerate}
The above procedure is known as MUSIC~\cite{joseMUSIC2013,kim2010compressive}.
Also, note that
we may use some other hyperspectral subspace identification algorithms, e.g., HySiMe \cite{Hysime}, to estimate the signal subspace matrix ${\bm U}_S$ from the noisy ${\bm Y}$.
MUSIC can in principle be used to perform dictionary-based semiblind HU.
However, because of its sensitivity to colored noise and modeling error that are usually present in real data,
it is used as a preprocessing algorithm for CSR (or other sparse regression methods) in practice.
Specifically, MUSIC is used to discard a large number of spectral samples that yield large residuals $\gamma_{\rm MUSIC}(k)$.
The remaining spectral samples then form a (much) smaller dictionary for CSR to operate.
Such a dictionary pruning procedure has been found to be able to improve the HU performance and speed up the process quite significantly---see \cite{joseMUSIC2013} for the detail.


\section{Proposed Approach}

The crucial assumption with dictionary-aided semiblind HU is that there is no spectrum mismatches; that is,
we can always find a dictionary sample that {\em exactly} matches an actual spectral signature in the scene; cf. Eq.~\eqref{eq:exact_match}.
As discussed in the Introduction, this may be not the case in reality.
In this section, we will propose a dictionary-aided semiblind HU that takes into account the presence of spectrum mismatches.

\subsection{Dictionary-Adjusted Nonconvex Sparsity-Encouraging Regression (DANSER)}

We assume the following spectrum mismatch model in place of \eqref{eq:exact_match}:
\begin{equation}
        {\bm d}_{k_n} = {\bm a}_n + {\bm \varepsilon}_n, \quad n=1,\ldots,N,
\label{eq:bk}
\end{equation}
for some ${\bm \varepsilon}_n \in \mathbb{R}^M$ that characterizes the mismatch between the presumed and actual spectra of each material.
Particularly, every spectral error ${\bm \varepsilon}_n$ is assumed to be bounded:
\[\| {\bm \varepsilon}_n \|_2\leq \delta,\quad n=1,\ldots,N,\]
for some $\delta > 0$.
Physically,
our model assumes that the dictionary still covers all the actual spectral signatures in the scene, but their ``best matched'' spectral samples in the dictionary are subject to certain perturbations.
Also, such perturbations do not go worse than $\delta^2$
in terms of magnitude.

Our rationale is to adjust the dictionary in the CSR formulation.
Specifically, we write
\[{\bm d}'_k = {\bm d}_k + {\bm e}_k,\quad k=1,\ldots,K,\]
where each ${\bm e}_k \in \mathbb{R}^M$ is a dictionary correction variable and we assume $\|{\bm e}_k\|_2\leq \delta$.
Following the CSR formulation in \eqref{eq:CSR},
we propose a new formulation as follows:
\begin{equation}
\begin{aligned}
\min_{{\bm D}'\in\mathbb{R}^{M\times K},~{\bm C}\in\mathbb{R}^{K\times L}}~&\frac{1}{2}\|{\bm Y}-{\bm D}'{\bm C}\|_F^2 + \lambda \|{\bm C}\|_{2,p}^p\\
{\rm s.t.}~&\|{\bm d}_k'-{\bm d}_k\|_2\leq \epsilon,\quad k=1,\ldots,K,\\
           &{\bm C}\geq{\bm 0},
\end{aligned}\label{eq:M-CSR}
\end{equation}
where $0<p< 1$, $\lambda >0$ and  $\epsilon>0$ are prespecified, and note that $\|{\bm C}\|^p_{2,p} = \sum_{i=1}^K \| {\bm c}^i \|_2^p$.
Comparing the original CSR formulation in \eqref{eq:CSR} and the above formulation, we see two differences.
First, Problem~\eqref{eq:M-CSR} adjusts the dictionary to attempt to neutralize the spectrum mismatches.
Second, Problem~\eqref{eq:M-CSR} employs a nonconvex row-sparsity promoting function $\|{\bm C}\|_{2,p}^p$.
The reason is that nonconvex $\ell_p$ quasi-norms may exhibit better sparsity promoting performance than the $\ell_1$-norm, as reported in the sparse optimization context \cite{chartrand2008restricted,Chartrand2008,shen2013exact},
and we endeavor to explore such an opportunity to improve sparse regression performance in the HU application.
The formulation in \eqref{eq:M-CSR} or its variants will be called dictionary-adjusted nonconvex sparsity-encouraging regression (DANSER) in the sequel.

\subsection{An Efficient Algorithm for DANSER}

Having expressed the DANSER formulation in the last subsection, we turn our attention to algorithm design for DANSER.
A simple approach to handle DANSER is to apply alternating optimization:
fix ${\bm D}'$ and
optimize Problem~\eqref{eq:M-CSR} with respect to (w.r.t.) ${\bm C}$ at one time,
fix ${\bm C}$ and
optimize Problem~\eqref{eq:M-CSR} w.r.t. ${\bm D}'$ at another time,
and repeat the above cycle until some stopping criterion holds.
While this approach is doable,
our algorithm design experience is that it can lead to a computationally expensive algorithm.
For instance, the optimization of Problem~\eqref{eq:M-CSR} w.r.t. ${\bm D}'$ involves joint adjustment of all the dictionary samples in an inseparable manner, which is computationally involved for large dictionary sizes.
Also, the nonconvex row-sparsity promoting function $\|{\bm C}\|_{2,p}^p$ used in Problem~\eqref{eq:M-CSR} introduces difficulties in the optimization of Problem~\eqref{eq:M-CSR} w.r.t. ${\bm C}$.

In view of the aforementioned issues, we formulate a modified version of Problem~\eqref{eq:M-CSR}:
\begin{equation}
\begin{aligned}
\min_{{\bm D}',{\bm H},{\bm C}}\quad&\frac{1}{2}\|{\bm Y}-{\bm H}{\bm C}\|_F^2 +\frac{\mu}{2}\|{\bm H}-{\bm D}'\|_F^2\\&\quad\quad\quad\quad\quad+ \lambda \sum_{k=1}^K\left(\|{\bm c}^k\|_{2}^2+\tau\right)^{p/2}\\
{\rm s.t.}\quad&\|{\bm d}_k'-{\bm d}_k\|_2\leq \epsilon,\quad k=1,\ldots,K,\\
           &{\bm C}\geq{\bm 0},
\end{aligned}\label{eq:DANSR}
\end{equation}
where $\mu,\tau>0$, and ${\bm H}$ is a slack variable.
In particular, it can be verified that if $\mu = +\infty$ and $\tau=0$, then Problem~\eqref{eq:DANSR} and Problem~\eqref{eq:M-CSR} are essentially the same.
It should be noted that we have applied the variable splitting technique in Problem~\eqref{eq:DANSR} (specifically, to the variable ${\bm C}$),
which is a commonly used trick in contexts such as image reconstruction~\cite{courant1943variational,wang2008new,xiao2010fast}.

The modified DANSER formulation in \eqref{eq:DANSR} can be handled in a low per-iteration complexity fashion.
To describe it, let us first
consider the following lemma \cite{fu2015joint,geman1992constrained,idier2001convex}:
\begin{Lemma}\label{lem:conjugate}
    Let $\phi_p(w) = \frac{2-p}{2}\left(\frac{2}{p}w \right)^{\frac{p}{p-2}}+\tau w$,
    where $0<p<2$, $\tau > 0$.
    The function $\phi_p(w)$ is strictly convex on $w\geq 0$.
		Also, $\phi_p(w)$ satisfies the following identity
		\begin{align*}
		  &\left(x^2+\tau\right)^{p/2} = \min_{w\geq 0}~w \cdot x^2+\phi_p(w)
		\end{align*}
		and the solution to the problem above is uniquely given by
		\begin{equation}
    w_{\rm opt} = \frac{p}{2}\left(x^2+\tau\right)^{\frac{p-2}{2}}.
		\label{eq:w_opt}
		\end{equation}
\end{Lemma}
%
%
%
By Lemma~\ref{lem:conjugate}, Problem~\eqref{eq:DANSR} can be equivalently expressed as
\begin{equation}
				\begin{aligned}
         \min_{{\bm H},{\bm C},{\bm D}',\{w_k\}}\quad&\frac{1}{2}\|{\bm Y}-{\bm H}{\bm C}\|_F^2+\frac{\mu}{2}\|{\bm H}-{\bm D}'\|_F^2\\&\quad+ \lambda\sum_{k=1}^K \left(w_k\left\|{\bm c}^k\right\|_2^2+\phi_p(w_k)\right)\\
				             {\rm s.t.}\quad&\| {\bm d}_{k}'-{\bm d}_{k}\|_2\leq \epsilon,~k=1,\ldots,K,\\
										            &{\bm C}\geq {\bm 0},\\
																&w_k\geq 0,~k=1,\ldots,K,
			\end{aligned}
			\label{eq:STLS2}
\end{equation}
Now, our strategy is to perform alternating optimization w.r.t. ${\bm H}$, ${\bm D}'$, $\{w_k\}$, ${\bm c}^1,\ldots,{\bm c}^K$.
As we will see soon, the merit of doing so is that every update admits a computationally light solution.

First, we examine the optimization w.r.t. ${\bm H}$.
One can easily see that the solution is
\begin{equation}\label{eq:sln_H}
	 {\bm H}:=(\mu{\bm D}'+{\bm Y}{\bm C}^T) \left({\bm C}{\bm C}^T+\mu{\bm I}\right)^{-1}.
\end{equation}
Second, the optimization w.r.t. ${\bm D}'$ is separable w.r.t. ${\bm d}_1', \ldots {\bm d}_K'$, i.e., for $k=1,\ldots,K$, we have
\begin{equation}
\begin{aligned}
	         \min_{{\bm d}_k'}~&\|{\bm d}_k'-{\bm h}_{k}\|_2^2\\
					 {\rm s.t.}~&\| {\bm d}_{k}'-{\bm d}_{k}\|_2\leq \epsilon.
\end{aligned}\label{eq:d_k_proj}
\end{equation}
Problem~\eqref{eq:d_k_proj} is a projection problem and the solution is
\begin{equation}
	      {\bm d}_{k}':=\begin{cases}    {\bm h}_k, &\quad \| {\bm h}_{k}-{\bm d}_{k}\|_2\leq \epsilon\\
						                           {\bm d}_{k}+\epsilon\frac{{\bm h}_k-{\bm d}_{k}}{\|{\bm h}_k-{\bm d}_{k}\|_2},&\quad \text{otherwise}       \end{cases}.
\end{equation}
Third, to check the solution w.r.t. ${\bm c}^k$, let us
first re-write the optimization w.r.t. ${\bf C}$ as
\begin{equation*}
			 \begin{aligned}
				  {\min_{{\bm C}}}~&\left\| \tilde{\bm Y}-\tilde{\bm H}{\bm C}\right\|_F^2\\
					 {\rm s.t.}~& {\bm C} \geq {\bm 0},
			 \end{aligned}
\end{equation*}
where
\[     \tilde{\bm Y}=\begin{bmatrix}\sqrt{\frac{1}{2}}{\bm Y}\\{\bm 0}\end{bmatrix},\quad   \tilde{\bm H}=\begin{bmatrix}\sqrt{\frac{1}{2}}{\bm H}\\ {\rm Diag}({\bm \theta})\end{bmatrix},        \]
and ${\bm \theta}:=[~\sqrt{{w}_1\lambda},\ldots,\sqrt{{w}_K\lambda}~]^T$.
Then, the subproblem w.r.t. ${\bm c}^k$ can be expressed as
\begin{equation}
			 \begin{aligned}
				  {\min_{{\bm c}^k}}~&\left\| \tilde{\bm Y}_k-\tilde{\bm h}_k{\bm c}^k\right\|_F^2\\
					 {\rm s.t.}~& {\bm c}^k\geq {\bm 0},
			 \end{aligned}\label{eq:sub_s}
\end{equation}
where
\[     \tilde{\bm Y}_k=\begin{bmatrix}\sqrt{\frac{1}{2}}{\bm Y}-\sum_{j\neq k}\sqrt{\frac{1}{2}}{\bm h}_j{\bm c}^j\\{\bm 0}\end{bmatrix},\quad   \tilde{\bm h}_k=\begin{bmatrix}\sqrt{\frac{1}{2}}{\bm h}_k\\\sqrt{{w}_k\lambda}{\bm f}_k\end{bmatrix},       \]
where ${\bm f}_k$ is the $k$th column of the $K\times K$ identity matrix.
Problem~\eqref{eq:sub_s} is known to have a simple solution \cite{cichocki2009fast,chemometrics1998least}, given by
\begin{align}
	          ({\bm c}^k)^T:
											&=\left(\frac{[\tilde{\bm Y}^T\tilde{\bm H}]_{:,k}-{\bm C}^T[\tilde{\bm H}^T\tilde{\bm H}]_{:,k} + ({\bm c}^k)^T[\tilde{\bm H}^T\tilde{\bm H}]_{k,k}  }{[\tilde{\bm H}^T\tilde{\bm H}]_{k,k}}\right)_{+}, \label{eq:s_3}
\end{align}
where $(x)_{+}=\max\{0,x\}$. 
Notice that using the update \eqref{eq:s_3} is desirable: the large matrix products $\tilde{\bm Y}^T\tilde{\bm H}$ and $\tilde{\bm H}^T\tilde{\bm H}$ both only need to be calculated once before updating ${\bm c}^1,\ldots,{\bm c}^K$.
Finally, by Lemma~\ref{lem:conjugate}, the solution w.r.t. $\{w_k\}$ is
\begin{equation}
	    w_k =  (p/2)(\|{\bm c}^k\|_2^2+\tau)^{(p-2)/2},\quad k=1,\ldots,K.
\end{equation}

The alternating optimization process described above is summarized in  Algorithm~\ref{Algo:newDANSER}, and we simply call it DANSER.
The DANSER algorithm has the following solution convergence guarantee.
\begin{Prop}\label{prop:convergence}
	    Every limit point of the solution sequence produced by DANSER (Algorithm~\ref{Algo:newDANSER}) is a stationary point of Problem~\eqref{eq:DANSR}.
\end{Prop}
The proof of the above proposition is relegated to Appendix~\ref{app:convergence}.
Proposition~\ref{prop:convergence} indicates that, although we have been dealing with Problem~\eqref{eq:DANSR} indirectly,
a stationary point of Problem~\eqref{eq:DANSR} may be expected.
Following Proposition~\ref{prop:convergence}, we can stop DANSER by checking the relative or absolute change of the solution ${\bm C}$.
Notice that since Problem~\eqref{eq:DANSR} is nonconvex, a good initialization would help DANSER converge to a better solution.
In practice, one can use
the CSR solution mentioned in Section~II.A
to initialize DANSER.

\begin{Remark}
By analyzing the per-iteration complexity of DANSER, one can verify that the complexities of many operations scale with $K$ (i.e., the size of the dictionary) or higher.
For example, to solve \eqref{eq:sln_H},  the operations ${\bm C}{\bm C}^T$ and ${\bm Y}{\bm C}^T$ cost ${\cal O}(K^2L)$ and ${\cal O}(MKL)$ flops, respectively;
and the matrix inversion requires ${\cal O}(K^3)$ flops.
Plus, although solving the problems w.r.t. ${\bm c}^k$ is easy, these procedures have to be repeated $K$ times at each iteration.
Practically, it is therefore motivated to use a dictionary with a smaller size, or, to prune the dictionary in advance.
However, due to the existence of spectral signature mismatches, directly applying MUSIC as in \cite{joseMUSIC2013} for this purpose is not appropriate any more.
To address this problem, a robust dictionary pruning method will be proposed in the next subsection.
\end{Remark}


\begin{algorithm}[!h]
\SetKwInOut{Input}{input}\SetKwInOut{Output}{output}
\SetKwRepeat{Repeat}{repeat}{until}

\Input{$(\lambda, \tau, p,\mu, \epsilon)$; ${\bm D}$; ${\bm C}$ (initialization); ${\bm Y}$. }

$w_k =  (p/2)(\|{\bm c}^k\|_2^2+\tau)^{(p-2)/2}$ for $k=1,\ldots,K$.

\Repeat{some stopping criterion is satisfied}{

\textbf{Unmixing}: construct ${\bm \theta}:=[~\sqrt{w_1\lambda},\ldots,\sqrt{w_k\lambda}~]^T$;
\[     \tilde{\bm Y}=\begin{bmatrix}\sqrt{\frac{1}{2}}{\bm Y}\\{\bm 0}\end{bmatrix},\quad   \tilde{\bm H}=\begin{bmatrix}\sqrt{\frac{1}{2}}{\bm H}\\ {\rm Diag}({\bm \theta})\end{bmatrix};        \]

 let ${\bm F}:=\tilde{\bm Y}^T\tilde{\bm H}$ and ${\bm G}:=\tilde{\bm H}^T\tilde{\bm H}$;

\For{$k=1:K$}{
update ${\bm c}^k$ by
\begin{align*}
	          {\bm c}^k:=\left(\frac{{\bm F}_{:,k}-{\bm C}^T{\bm G}_{:,k} + ({\bm c}^k)^T{\bm G}_{k,k}  }{{\bm G}_{k,k}}\right)_{+}. \label{eq:s_3}
\end{align*}
}

\textbf{Dictionary Adjusting}: update ${\bm H}$ by

\begin{equation*}
	 {\bm H}:= (\mu{\bm D}'+{\bm Y}{\bm C}^T)\left({\bm C}{\bm C}^T+\mu{\bm I}\right)^{-1}.
\end{equation*}

\textbf{Update Slack Variable}:

\For{$k=1:K$}{
\begin{equation*}
	      {\bm d}_{k}':=\begin{cases}    {\bm h}_k, &\quad \| {\bm h}_{k}-{\bm d}_{k}\|_2\leq \epsilon\\
						                           {\bm d}_{k}+\epsilon\frac{{\bm h}_k-{\bm d}_{k}}{\|{\bm h}_k-{\bm d}_{k}\|_2},&\quad {\rm o.w.}       \end{cases};											
\end{equation*}
}

\textbf{Reweighting}: update $\{w_k\}$ by
\begin{equation*}
				 w_k =  (p/2)(\|{\bm c}^k\|_2^2+\tau)^{(p-2)/2},\quad k=1,\ldots,K.
\end{equation*}						

}

\Output{${\bm C}$.}

\caption{DANSER}\label{Algo:newDANSER}
\end{algorithm}

\subsection{Robust MUSIC for Dictionary Pruning}

Consider the MUSIC procedure back in Section II.B.
In particular, recall that the metric
\begin{align} \label{eq:MUSIC2}
                             \gamma_{\rm MUSIC}(k) = \frac{{\bm d}_k^T{\bm P}^{\perp}_{{\bm U}_S}{\bm d}_k}{\|{\bm d}_k\|_2^2}
                \end{align}
should yield a small value when ${\bm d}_k$ exactly matches an actual spectral signature in the scene,
and this property has been used as the way to prune the dictionary in the MUSIC procedure.
Now, in the presence of dictionary mismatches, we propose to replace \eqref{eq:MUSIC2} by the following robust MUSIC (RMUSIC) metric
\begin{subequations}\label{eq：robustMUSIC}
                \begin{align}
                             \gamma_{\rm RMUSIC}(k) = \min_{{\bm \xi}\in\mathbb{R}^M}~&\frac{({\bm d}_k-{\bm \xi})^T{\bm P}^{\perp}_{{\bm U}_S}({\bm d}_k-{\bm \xi})}{\|  {\bm d}_k -{\bm \xi}\|_2^2}\\
                                                        {\rm s.t.}~&\|{\bm \xi}\|_2\leq {\epsilon} \label{eq:ball_constraint},
                \end{align}
\end{subequations}
where $\epsilon > 0$ is prespecified.
The idea is the same as the DANSER development in the above subsections---adjust the dictionary to find a better match, this time in a subspace sense.

The key issue with realizing RMUSIC lies in solving Problem~\eqref{eq：robustMUSIC}.
Problem~\eqref{eq：robustMUSIC} is a single-ratio fractional quadratic program,
which is quasi-convex and can be solved, e.g., by the Dinkelbach algorithm or its variants \cite{zhang2008quadratic,dinkelbach1967nonlinear}.
While this means that we can implement RMUSIC by applying some existing optimization algorithms,
we have to solve $K$ such quasi-convex problems---which is still inefficient for large $K$.
However, by carefully examining the problem structure, we find that this particular problem can be solved quite easily.
To see this, let us re-express $\gamma_{\rm RMUSIC}(k)$ as
\begin{equation}\label{eq:11}
\begin{aligned}
      \gamma_{\rm RMUSIC}(k)&=\min_{\|{\bm \xi}\|_2\leq\epsilon}~\frac{\left\|{\bm P}^{\perp}_{{\bm U}_S}({\bm d}_k-{\bm \xi})\right\|_2^2}{\|{\bm P}^{\perp}_{{\bm U}_S}({\bm d}_k-{\bm \xi})\|_2^2+\|{\bm P}_{{\bm U}_S}({\bm d}_k-{\bm \xi})\|_2^2}  \\
                                    &= \min_{\|{\bm \xi}\|_2\leq\epsilon} ~\frac{\eta_k^2({\bm \xi})}{\eta_k^2({\bm \xi})+1},
\end{aligned}
\end{equation}
where ${\bm P}_{{\bm U}_S}={\bm U}_S{\bm U}_S^T$ denotes the orthogonal projector of ${\bm U}_S$, and
\begin{equation}
    \eta_k({\bm \xi}) = \frac{\left\|{\bm P}^{\perp}_{{\bm U}_S}({\bm a}_k-{\bm \xi})\right\|_2}{\left\|{\bm P}_{{\bm U}_S}({\bm a}_k-{\bm \xi})\right\|_2}.
\label{eq:eta}
\end{equation}
Since the objective function of \eqref{eq:11} is a monotone increasing function of $\eta^2({\bm \xi})\in[0,\infty)$, computing $\gamma_{\rm RMUSIC}(k)$
is the same as finding the minimal value of $\eta_k({\bm \xi})$ subject to $\|{\bm \xi}\|_2\leq\epsilon$.
Let us denote
\begin{equation}\label{eq:eta_prob}
\begin{aligned}
      \eta_k^\star = \min_{\|{\bm \xi}\|_2\leq\epsilon}~ \frac{\left\|{\bm P}^{\perp}_{{\bm U}_S}({\bm a}_k-{\bm \xi})\right\|_2}{\left\|{\bm P}_{{\bm U}_S}({\bm a}_k-{\bm \xi})\right\|_2}.
\end{aligned}
\end{equation}
We show that
\begin{Prop}\label{prop:etastar}
The optimal value of Problem~\eqref{eq:eta_prob} can be found by solving a single-variable problem
\begin{equation}
         \eta_k^\star=\min_{0\leq \theta \leq \epsilon}~\frac{\left| \|{\bm P}^{\perp}_{{\bm U}_S}{\bm d}_k\|_2 - \theta \right|}{\|{\bm P}_{{\bm U}_S}{\bm d}_k\|_2 + \sqrt{\epsilon^2-\theta^2}}. \label{eq:etastar}
\end{equation}
\end{Prop}
The proof of Proposition~\ref{prop:etastar} is relegated to Appendix~\ref{app:rmusic}.
The message revealed here is quite intriguing---the originally quasi-convex problem can be recast into a simple single-variable problem that can be easily solved, e.g., by grid search or bisection.
Practically, this means that the RMUSIC strategy can be implemented quite efficiently.

As in the previous MUSIC work~\cite{joseMUSIC2013}, we use RMUSIC to perform dictionary pruning for DANSER.
Specifically, we use RMUSIC to select a number of $\tilde{K}$ ($\tilde{K}<K$) spectral samples from ${\bm D}$,
form a size-$\tilde{K}$ dictionary, denoted by $\tilde{\bm D}$ here,
and then use $\tilde{\bm D}$ as a pruned dictionary to run DANSER.
We summarize this procedure in Algorithm~\ref{Algo:RMUSIC-DANSER},
and we call the procedure \emph{RMUSIC-DANSER}.
\begin{algorithm}[!h]
\SetKwInOut{Input}{input}\SetKwInOut{Output}{output}
\SetKwRepeat{Repeat}{repeat}{until}

\Input{${\bm Y}$; ${\bm D}$; $\epsilon$; $\tilde{K}$. }

apply HySiMe \cite{Hysime} on ${\bm Y}$ to obtain ${\bm U}_S$;

\For{$k=1:K$}{
$ \eta_k^\star=\min_{0\leq \theta \leq \epsilon}~\frac{\left| \|{\bm P}^{\perp}_{{\bm U}_S}{\bm d}_k\|_2 - \theta \right|}{\|{\bm P}_{{\bm U}_S}{\bm d}_k\|_2 + \sqrt{\epsilon^2-\theta^2}}$;

$\gamma_{\rm RMUSIC}(k)=\frac{(\eta^\star_k)^2}{(\eta^\star_k)^2 + 1}$;
}

determine $\hat{\Lambda}=\{  \hat{k}_1,\ldots, \hat{k}_{\tilde{K}}  \}$ such that ${\gamma}_{\rm RMUSIC}(i)<{\gamma}_{\rm RMUSIC}(j)$ for any $i\in\hat{\Lambda}$ and $j\notin\hat{\Lambda}$;

determine
$\tilde{\bm D}= \left[ {\bm d}_{\hat{k}_1}, {\bm d}_{\hat{k}_2}, \ldots, {\bm d}_{\hat{k}_{\tilde{K}}} \right]$;

feed $\tilde{\bm D}$ and ${\bm Y}$ to DANSER (Algorithm~\ref{Algo:newDANSER});

\Output{${\bm C}$.}

\caption{RMUSIC-DANSER}\label{Algo:RMUSIC-DANSER}
\end{algorithm}

\section{Computer Simulations}\label{sec:simulations}
In this section, we use synthetic hyperspectral images to show the effectiveness of the proposed approach.
In our simulations, the ground-truth spectra are randomly selected from a subset of the U.S.G.S. library that has $332$ spectral signatures.
The abundances are generated following the uniform Dirichlet distribution.
Throughout this section, we set the number of pixels to be $L=5000$.
The `available dictionary', ${\bm D}$, is formed by the same subset of spectra, but a perturbation (i.e., ${\bm e}_k$ for $k=1,\ldots,K$) is intentionally added to each spectrum.
To quantify the `mismatch level' of the available dictionary, we define the \emph{dictionary to modeling error ratio} (DMER) as follows:
\[{\rm DMER} ({\rm dB}) = 10\log_{10}\left({\|{\bm d}_{k^\star}\|_2^2}/{\delta^2}\right),\]
where $k^\star = \arg\min_{k=1,\ldots,K}~\|{\bm d}_k\|_2$ and $\delta = \max_{k=1,\ldots,K}\|{\bm e}_k\|_2$.
The perturbation term ${\bm e}_k$ follows the zero-mean i.i.d. Gaussian distribution and is scaled to satisfy DMER specifications.
We also define the signal-to-noise ratio (SNR) by
${\rm SNR}=\frac{\sum_{\ell=1}^L\|{\bm A}{\bm s}[\ell]\|_2^2}{ML\sigma^2}$ to quantify the noise level, where $\sigma^2$ denotes the variance of the additive noise, which is also assumed to be zero-mean i.i.d. Gaussian.
The choice of the parameter $\epsilon$ is as follows
\[\epsilon = \frac{1-\alpha}{1+\alpha}\|{\bm d}_{k^\star}\|_2, \]
where $\alpha\in[0,1]$ is given.
The parameter $\alpha$ controls the correlation between the RMUSIC/DANSER-resulted dictionary member ${\bm d}_{k^\star}-{\bm \xi}$ and the original one.
Specifically, under $\|{\bm \xi}\|_2\leq \epsilon$, it can be shown that the above choice of $\epsilon$ leads to
$ \frac{({\bm d}_{k^\star}-{\bm \xi})^T{\bm d}_{k^\star}}{\|{\bm d}_{k^\star}-{\bm \xi}\|_2\|{\bm d}_{k^\star}\|_2}\geq \alpha$.

\begin{figure}
	\centering
	\psfrag{residual}{residue}
		\includegraphics[width=8cm]{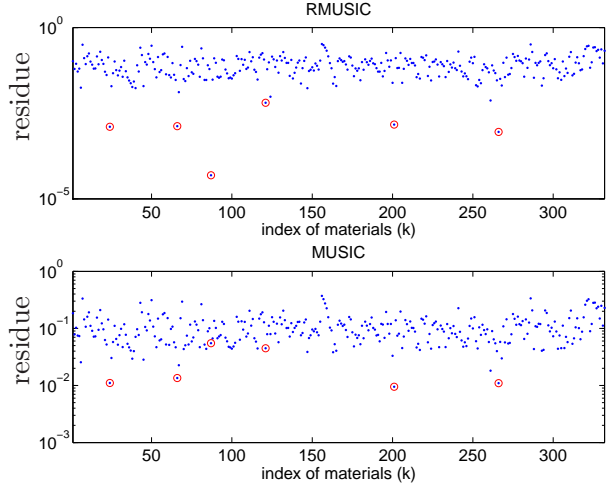}
	\caption{The projection residues of MUSIC and RMUSIC.}
	\label{fig:toy_rmusic}
\end{figure}

\begin{figure}
	\centering
	  \psfrag{ck}{$\|{\bm c}^k\|_2$}
		\includegraphics[width=8cm]{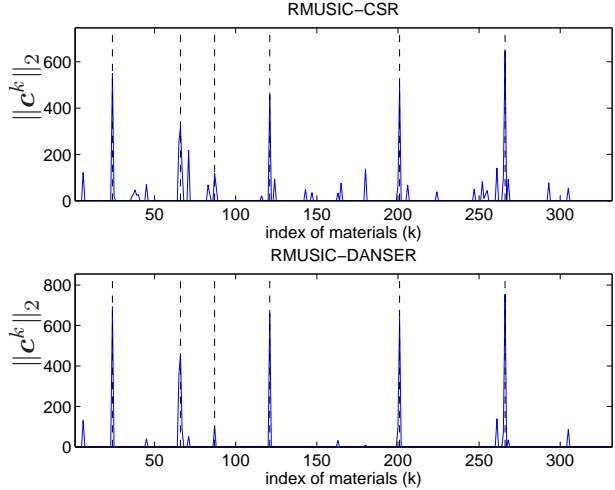}
	\caption{The 2-norms of ${\bm c}^k$'s of CSR and DANSER. The black dash lines correspond to the indices of the ground-truth materials' spectra in the dictionary;
	for the RMUSIC-pruned spectra, we set $\|{\bm c}^k\|_2=0$.}
	\label{fig:toy_danser}
\end{figure}

Figs.~\ref{fig:toy_rmusic}-\ref{fig:toy_danser} show an illustrative example.
Fig.~\ref{fig:toy_rmusic} shows the residues of applying MUSIC and RMUSIC to prune the dictionary ${\bm D}$.
Here, we randomly pick $N=6$ spectra as the ground-truth materials, and then use the described dictionary ${\bm D}$ to observe the performance of MUSIC and RMUSIC.
The parameter of RMUSIC is set to be $\alpha = 0.85$, and we set ${\rm DMER}=20$dB and ${\rm SNR}=35$dB in this case.
We see that MUSIC has difficulty in distinguishing several ground-truth spectra from the other dictionary members (to be precise, the third and the fourth materials' spectra), but
RMUSIC can clearly differentiate the same spectra from the irrelevant spectra.
Fig.~\ref{fig:toy_danser} compares the unmixing performance of DANSER and CSR using the same case, where the pruned dictionary with $40$ spectra is obtained by RMUSIC.
Here, the CSR part is performed by the CLSUnSAL algorithm \cite{Iordache2012collaborative}, which is considered as a state-of-the-art.
For DANSER, we set $p=0.5$, $\lambda = 0.04$ for this case. For CSR, the regularization parameter is $\lambda=0.005$.
In this example and the forthcoming simulations and real experiment, we feed the solution of CSR to DANSER as initialization.
We see that RMUSIC-DANSER yields much row-sparser ${\bm C}$ than that of RMUSIC-CSR, and all of the desired spectra have been successively identified by DANSER.

In the following, we use Monte Carlo simulations to evaluate the performance of the proposed algorithms.
Two performance discriminators will be used throughout this section.
First, to measure the dictionary pruning performance, we define the
following detection probability
\begin{equation*}
    {\rm Pr}\left\{\Lambda\subset \hat{\Lambda}\right\}
\label{eq:prob}
\end{equation*}
where ${\Lambda} = \{ k_1, \ldots, k_N \}$ denotes the index set that indicates the ground-true spectra, and $\hat{\Lambda} \subseteq \{1,\ldots, K \}$ denotes an index selection subset outputted by a dictionary pruning algorithm.
Also, we will use $\tilde{K}$ to denote the size of the pruned dictionary.
Second, to measure the unmixing performance, we calculate the following the signal to reconstruction error (SRE) \cite{Iordache2011,Iordache2012TV,Iordache2012collaborative}:
\[{\rm SRE(dB)} =10\log_{10}\left(\frac{\|{\bm S}\|_F^2}{\left\| {\bm C} - \hat{\bm C}\right\|_F^2}\right),\]
where
${\bm C}$ is the true row-sparse abundance matrix (see \eqref{eq:LMM_blk} in Section II.A),
and $\hat{\bm C}$ is the output of an unmixing algorithm.

\begin{figure}
	\centering
		\includegraphics[width=8cm]{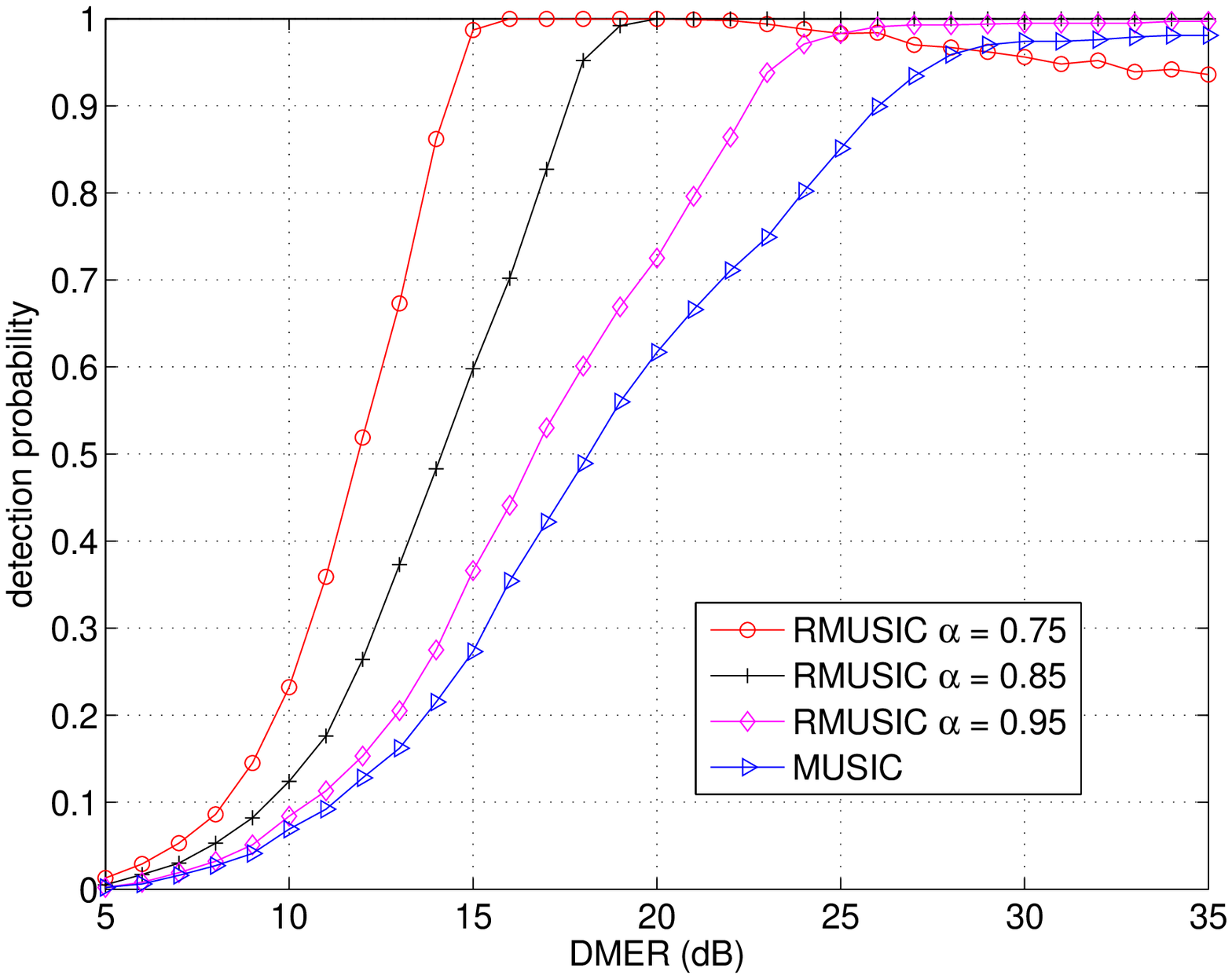}
	\caption{The detection probabilities of RMUSIC/MUSIC under various DMERs and different $\alpha$'s. SNR$=35$dB; $N=8$; the pruned dictionary size is $\tilde{K}=40$; the original dictionary size $K=332$.}
	\label{fig:rmusic_alpha}
\end{figure}

In Fig.~\ref{fig:rmusic_alpha}, we show the index set detection probabilities of MUSIC and RMUSIC under various DMERs.
In each trial, $N=8$ materials are randomly picked.
The ${\rm SNR}$ in this simulation is set to be $35$dB, and $\tilde{K}=40$ is employed.
The results are averaged from $1000$ trials.
One can see that MUSIC is sensitive to dictionary mismatches
even under high DMERs, and MUSIC is not able to identify all the true materials from the dictionary.
Generally, using RMUSIC with $\alpha=0.85$ and $0.95$ both yield much better detection probabilities than MUSIC under all DMERs.
Interestingly, one can see that RMUSIC with $\alpha=0.75$ admits very good detection probabilities when DMER$\leq 20$dB;
however, when the DMER is higher, using a small $\alpha$ leads to a slight performance degradation.
The reason is that a smaller $\alpha$ implies that one is allowed to adjust ${\bm d}_k$'s more significantly in RMUSIC.
Hence, several similar ${\bm d}_k$'s may be confused with each other.
This observation suggests that a more conservative choice of $\alpha$ should be safer for implementing RMUSIC in practice.

\begin{figure}
	\centering
		\includegraphics[width=8cm]{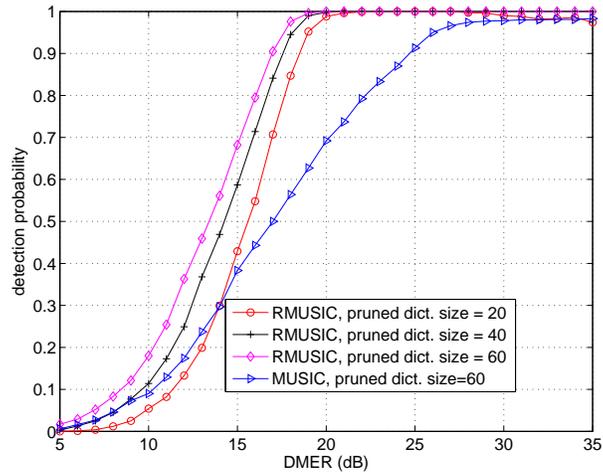}
	\caption{The detection probabilities of RMUSIC/MUSIC under various DMERs and different $\tilde{K}$'s (size of the pruned dictionary). $\alpha=0.85$; $N=8$; the original dictionary size $K=332$; SNR$=35$dB.}
	\label{fig:rmusic_Ktilde}
\end{figure}

Fig.~\ref{fig:rmusic_Ktilde} shows the detection probabilities of RMUSIC and MUSIC under different $\tilde{K}$'s (the size of the pruned dictionary).
Setting $\tilde{K}$ to be small may be easier for the sparse regression stage, but is considered more aggressive---some spectra corresponding to the ground-truth materials may also be discarded.
We see that when DMER$\geq 15$dB, RMUSIC with $\tilde{K}=20$ yields higher detection probabilities than that of MUSIC with $\tilde{K}=60$,
and that RMUSIC with a larger $\tilde{K}$ has a better detection performance.

\begin{figure}
	\centering
		\includegraphics[width=8cm]{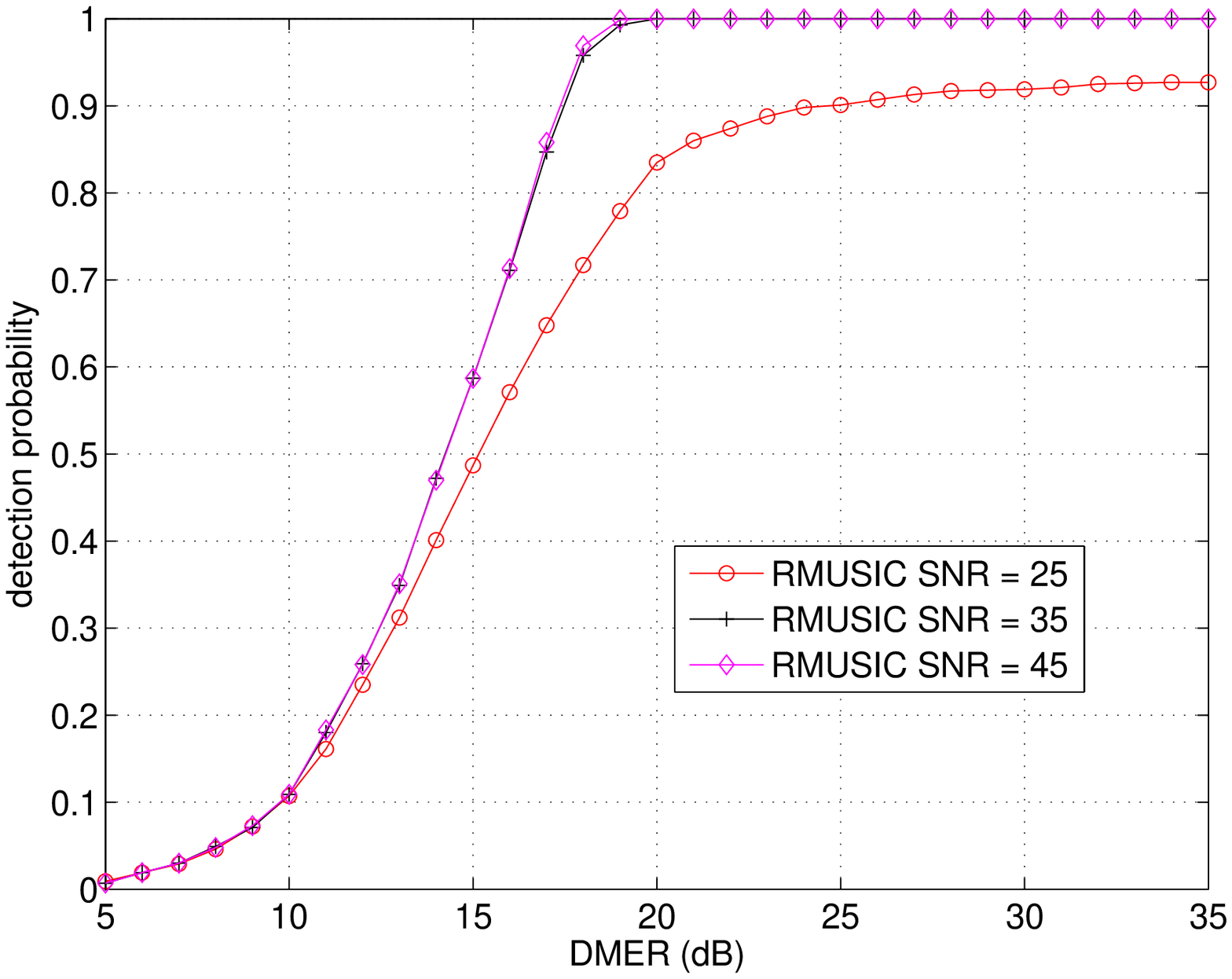}
	\caption{The detection probabilities of RMUSIC under various DMERs and different SNRs. $\alpha=0.85$; $N=8$; the pruned dictionary size is $\tilde{K}=40$; the original dictionary size $K=332$.}
	\label{fig:rmusic_SNR}
\end{figure}

\begin{figure}
	\centering
		\includegraphics[width=8cm]{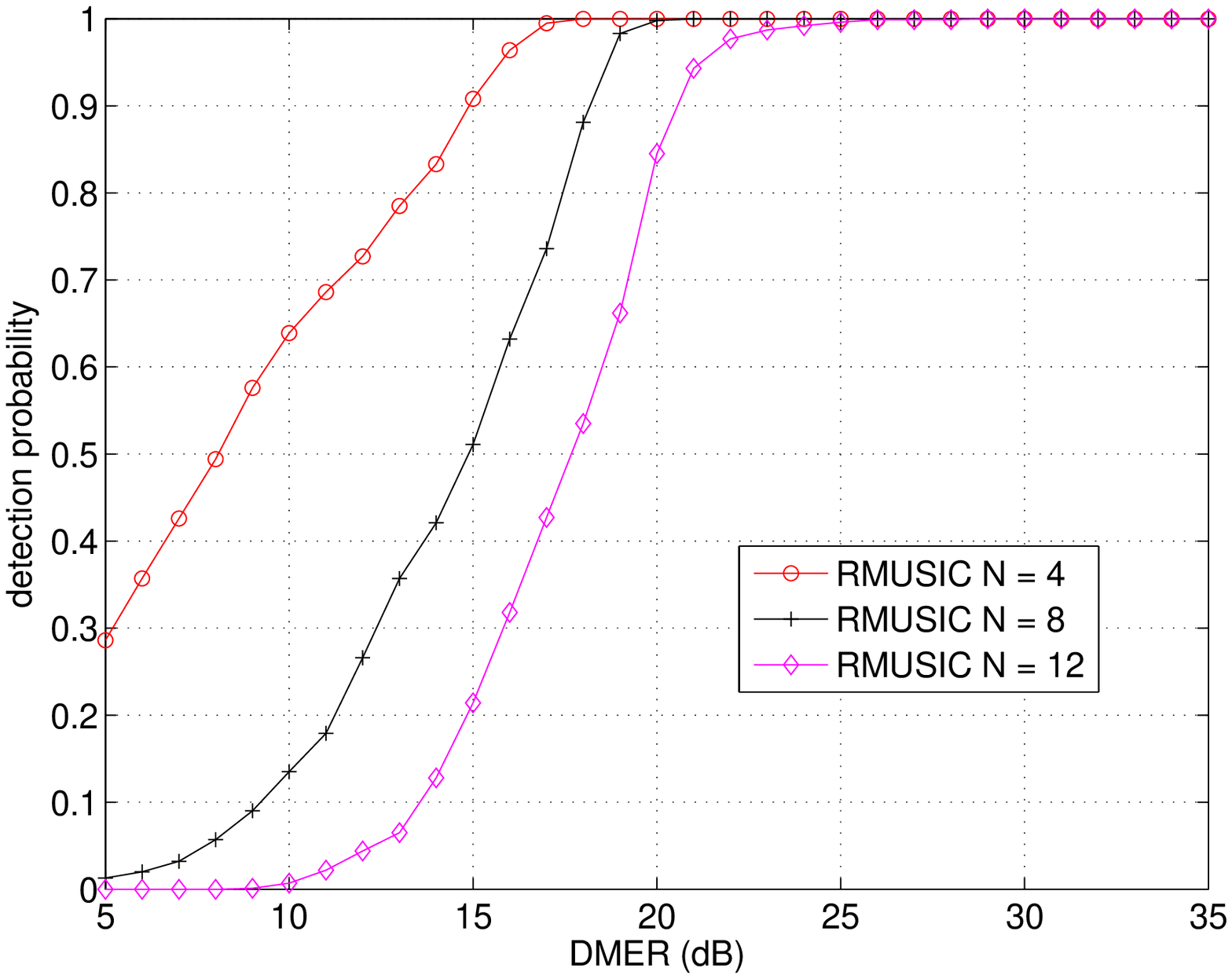}
	\caption{The detection probabilities of RMUSIC under various DMERs and different $N$'s. $\alpha=0.85$; $N=8$; the pruned dictionary size is $\tilde{K}=60$; the original dictionary size $K=332$; SNR$=35$dB.}
	\label{fig:rmusic_N}
\end{figure}

Fig.~\ref{fig:rmusic_SNR} and Fig.~\ref{fig:rmusic_N} show the performance of RMUSIC under various SNRs and underlying ground-truth materials, respectively.
From these figures,
one can see how this algorithm is scaled by different parameters.

\begin{figure}
	\centering
		\includegraphics[width=8cm]{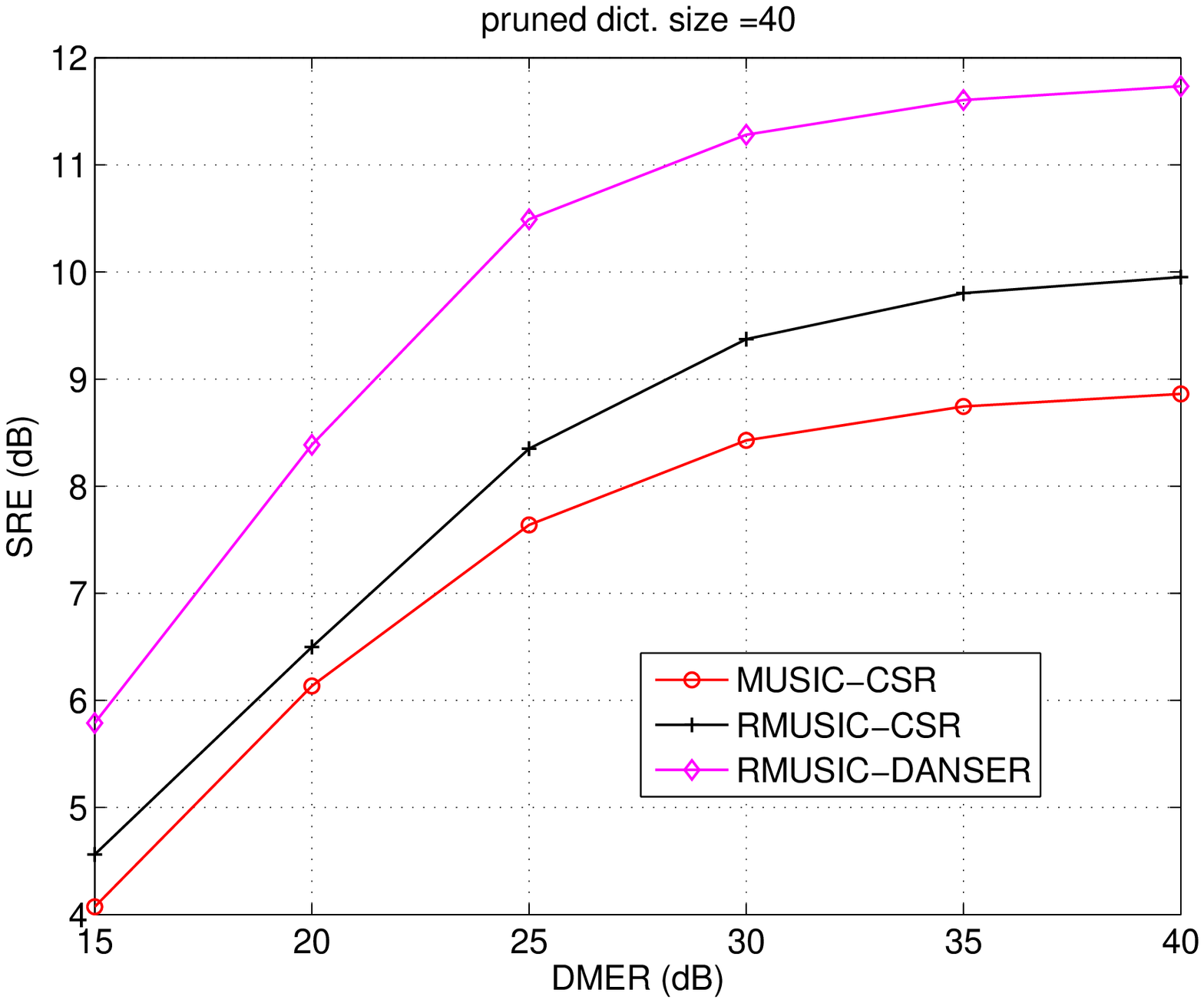}
	\caption{The SREs of the algorithms under different DMERs. $\alpha=0.85$; $N=8$; the pruned dictionary size is $\tilde{K}=40$; the original dictionary size $K=332$; SNR$=35$dB.}
	\label{fig:danser_dmer}
\end{figure}

Beginning from Fig.~\ref{fig:danser_dmer}, we show the SRE performance of the CSR-based HU algorithms.
Specifically, we compare the SREs yielded by the proposed RMUSIC-DANSER and by MUSIC-CSR \cite{joseMUSIC2013}.
We also benchmark our algorithm using RMUSIC-CSR for fairness, since we now have seen that RMUSIC yields much better dictionary pruning performance.
In all the following simulations, we fix $p=0.5$, $\mu=10^5$, $\tau=10^{-6}$ for DANSER, no matter how the simulation settings change;
the sparsity-controlling parameter $\lambda$ for DANSER and CSR are also fixed to be $0.5$ and $0.1$ except specified.
We stop DANSER if $\|{\bm C}^{(i)}-{\bm C}^{(i-1)}\|_F\leq 10^{-5}$, where ${\bm C}^{(i)}$ denotes the solution at iteration $i$, or if the number of iterations reaches $5000$.
The results in all the following figures of this section are averaged from $50$ independent trials.

Fig.~\ref{fig:danser_dmer} shows the SREs of the algorithms under different DMERs. We see that under all DMERs, RMUSIC-DANSER yields the highest SREs.
We see that RMUSIC-CSR also consistently yields better SRE performance than that of MUSIC-CSR---this suggests that RMUSIC itself can help improve the sparse unmixing performance.
The runtime performance of DANSER and CSR (i.e., CLSUnSAL) is shown in Table~\ref{tab:danser_dmer} as a reference.
We see that DANSER requires more time to converge compared to CSR, since it also adjusts the dictionary during its updates.
Also, when the DMER gets higher, the convergence speed of DANSER improves by $1/3$. This intuitively suggests that DANSER does put much effort on adjusting the dictionary (i.e., updating ${\bm H}$)
when the DMER is low.

\begin{table}[htbp]
  \centering
  \caption{The runtimes (sec.) of DANSER and CSR under various DMERs. $\alpha=0.85$; $N=8$; the pruned dictionary size is $\tilde{K}=40$; the original dictionary size $K=332$; SNR$=35$dB.}
	\resizebox{8.5cm}{!}{\huge
    \begin{tabular}{c|c|c|c|c|c|c}
    \hline
    \hline
    \multirow{2}[4]{*}{Algorithm} & \multicolumn{6}{c}{DMER (dB)} \\
\cline{2-7}          & 15    & 20    & 25    & 30    & 35    & 40 \\
    \hline
    \hline
    DANSER &  15.9205 &  16.2639 &  13.0023 &  11.0784  & 10.7352 &  9.9090 \\
    \hline
    CSR   &   0.8687  &  0.9123  &  1.3611  &  1.3472   & 1.6298 &   1.4699\\
    \hline
    \hline
    \end{tabular}}%
  \label{tab:danser_dmer}%
\end{table}%

\begin{figure}
	\centering
		\includegraphics[width=8cm]{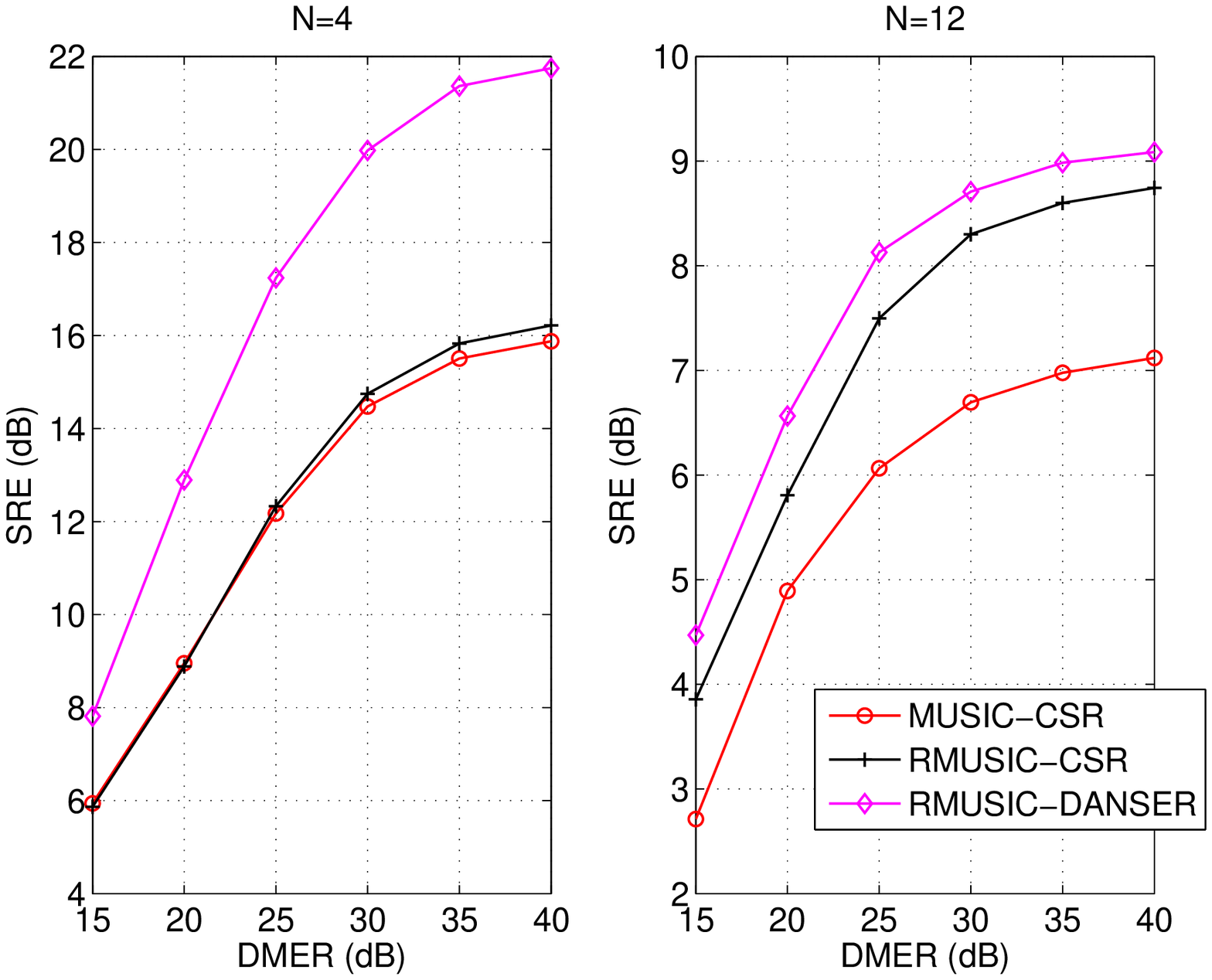}
	\caption{The SREs of the algorithms under different $N$'s. $\alpha=0.85$; $N=8$; the pruned dictionary size is $\tilde{K}=40$; the original dictionary size $K=332$; SNR$=35$dB.}
	\label{fig:danser_N}
\end{figure}

\begin{figure}
	\centering
		\includegraphics[width=8cm]{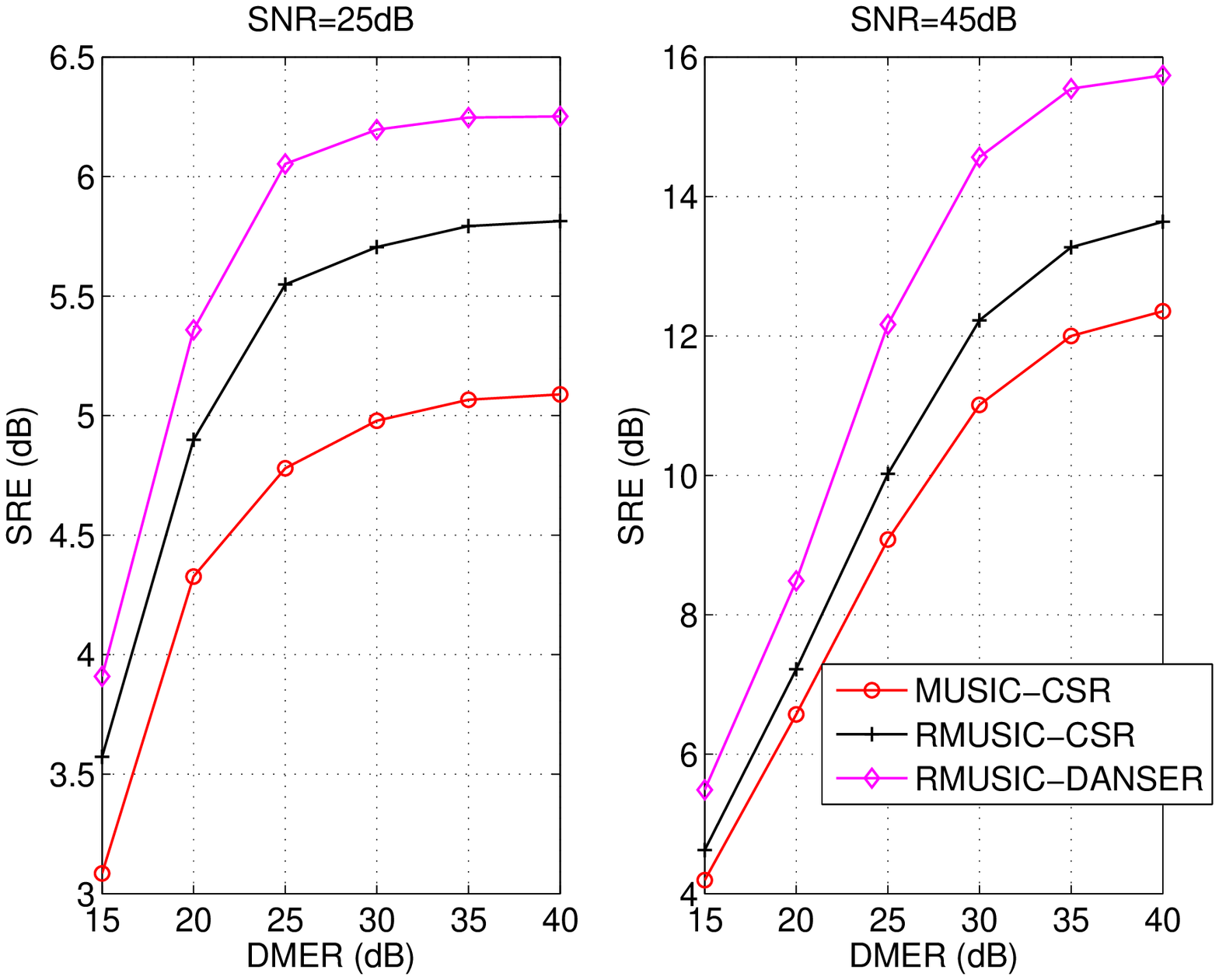}
	\caption{The SREs of the algorithms under different SNRs. $\alpha=0.85$; $N=8$; the pruned dictionary size is $\tilde{K}=40$; the original dictionary size $K=332$.}
	\label{fig:danser_snr}
\end{figure}

Fig.~\ref{fig:danser_N} and Fig.~\ref{fig:danser_snr} show the performance of the algorithms under different number of materials and SNRs, respectively.
We see that the results are similar to that in Fig.~\ref{fig:danser_dmer} --- the SRE performance of RMUSIC-DANSER is consistently higher than the other two under comparison.
Notice that for the SNR$=25$dB case, we change $\lambda$ of DANSER and CSR to be $1$ and $0.5$, respectively, to accommodate the situation where the data is more severely corrupted.

\begin{figure}
	\centering
		\includegraphics[width=8cm]{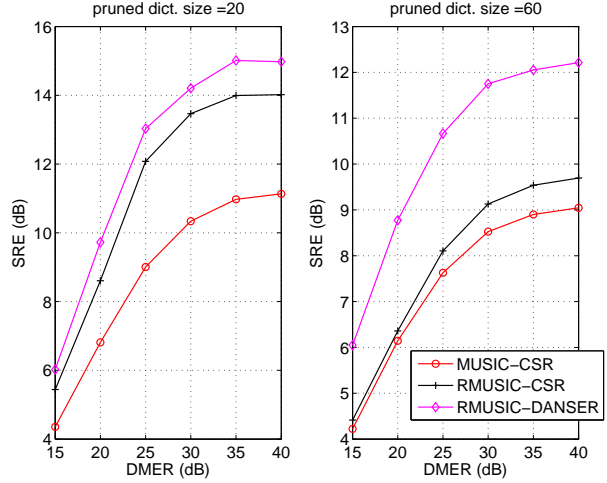}
	\caption{The SREs of the algorithms under different $\tilde{K}$s. $\alpha=0.85$; $N=8$; the original dictionary size $K=332$; SNR$=35$dB.}
	\label{fig:danser_Nu}
\end{figure}

Fig.~\ref{fig:danser_Nu} shows the SREs of the algorithms under different values of $\tilde{K}$.
An interesting observation is that using $\tilde{K}=20$ yields much better unmixing performance than using $\tilde{K}=60$.
This results may shed some light on choosing $\tilde{K}$ in practice - using a large $\tilde{K}$ may safely capture all the true materials in the pruned dictionary,
but it may also degrade the unmixing performance since the sparse regression-type algorithms are in general in favor of smaller $\tilde{K}$.

\section{Real Data Experiment}
\begin{figure}
	\centering
		\includegraphics[width=6cm]{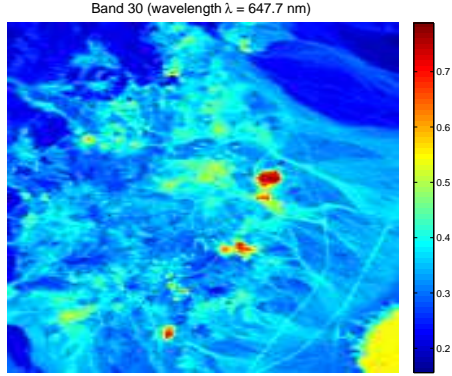}
	\caption{Band 30 (wavelength $\lambda = 647.7$ nm) of the subimage of AVIRIS
Cuprite Nevada data set that is used in the experiment of this section.}
	\label{fig:raw_image}
\end{figure}

In this section, we test the algorithms on the famous AVIRIS Cuprite data set which was captured in Nevada, 1997 (see \url{http://aviris.jpl.nasa.gov/html/aviris.freedata.html}).
This data set has been studied for years and the abundance maps of several prominent materials are well recognized.
The scene originally has 224 spectral bands between 0.4 and
2.5 µm, with nominal spectral resolution of 10 nm. Low SNR bands, i.e., bands 1--2, 105--115, 150--170, and 223--224, have been
removed, resulting a total of 188 spectral bands.
We take a subimage of the whole data set, which consists of 250$\times$191 pixels; see Fig.~\ref{fig:raw_image} for this subimage at spectral band $30$.
The dictionary that we use here is also the same as the one that has been used in the simulations.
It has been noticed that there are calibration mismatches between the real
image spectra of this scene and the spectra available in the U.S.G.S. library \cite{Iordache2011,Iordache2012TV,Iordache2012collaborative}.
Hence, this dataset is suitable for verifying our proposed algorithm.


\begin{figure}
	\centering
	\psfrag{MUSIC}{\tiny residue}
	\psfrag{RMUSIC}{\tiny residue}
		\includegraphics[width=8cm]{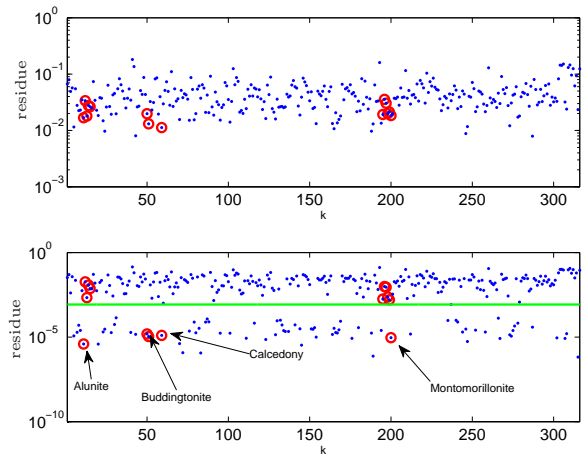}
	\caption{The MUSIC (top) and RMUSIC (bottom) residues of the real data.}
	\label{fig:real_data_subspace}
\end{figure}

We first apply RMUSIC and MUSIC on this data set.
We adopt the following way to qualitatively evaluate the performance:
From the previous studies in \cite{Iordache2011,Iordache2012TV,Iordache2012collaborative}, we know that
the prominent materials are \emph{Alunite}, \emph{Buddingtonite}, \emph{Chalcedony}, and \emph{Mmontmorillonite}.
Fig.~\ref{fig:real_data_subspace}
shows the residues obtained by applying MUSIC (top) and RMUSIC (buttom).
For RMUSIC, we set $\alpha = 0.85$.
The red circles correspond to the library members associated with Alunite, Buddingtonite, Chalcedony, and Mmontmorillonite.
We see that the residues given by RMUSIC can be clearly separated into two groups,
and the group with smaller residues include the spectra of the materials that we wish to identify.
We should emphasize that the situation in Fig.~\ref{fig:real_data_subspace} (bottom) is desirable in practice:
since the residues associated with the spectra are clearly divided into to two clusters, it is easy to decide which spectra should be kept in the pruned dictionary.
In this experiment,
we simply keep the spectra below the green line, which is drawn by visual inspection,
and this results in a pruned dictionary with $\tilde{K}=79$ spectra.
Compared to the original size $K=332$, RMUSIC successfully reduces the dictionary size by 75\% while preserving the spectra associated with the prominent materials.

\begin{figure*}
\begin{minipage}[t]{0.33\linewidth}
\subfigure{
\centering
\includegraphics[height=1.35in,width=1.7in]{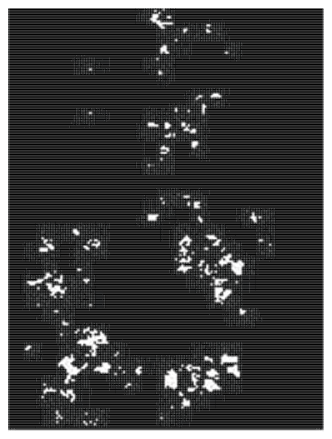}
}
\end{minipage}
\begin{minipage}[t]{0.33\linewidth}
\subfigure{
\centering
\includegraphics[width=1.7in]{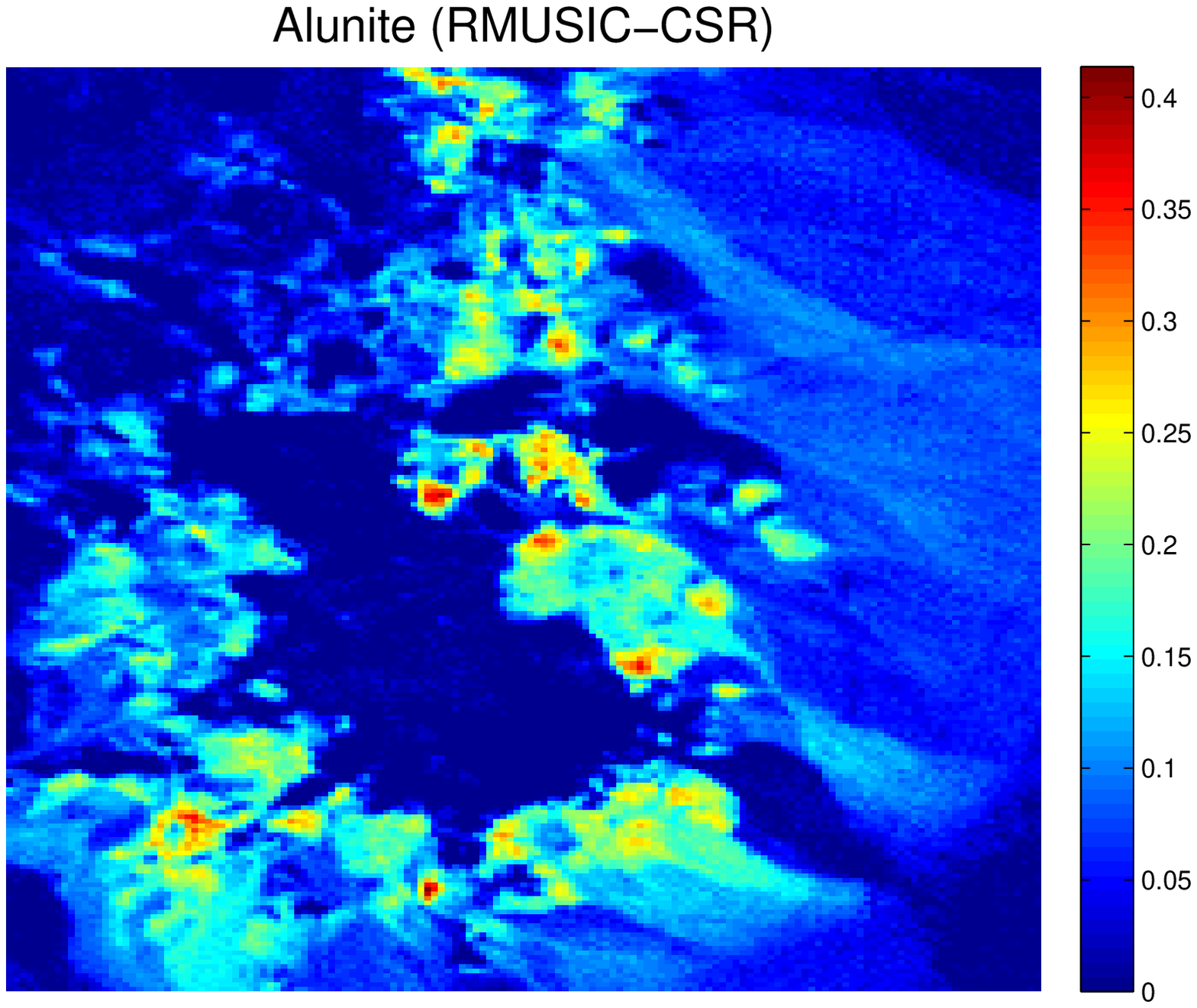}
}
\end{minipage}%
\begin{minipage}[t]{0.33\linewidth}
\subfigure{
\centering
\includegraphics[width=1.7in]{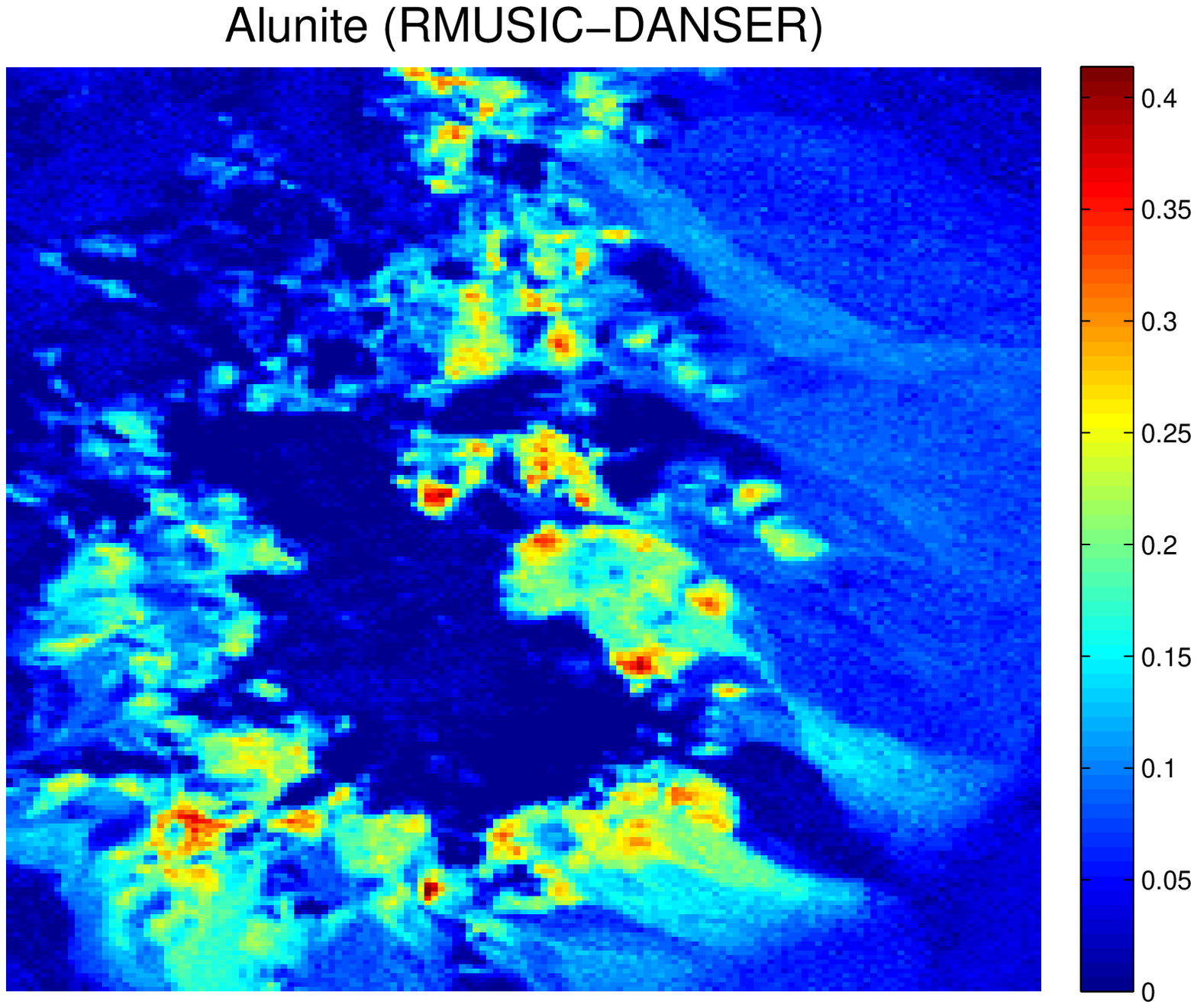}
}
\end{minipage}
\caption{The U.S.G.S. Tetracorder abundance map (left) and the estimated abundance map of the Alunite by RMUSIC-CSR and RMUSIC-DANSER, respectively.}\label{fig:abundance_1}
\end{figure*}

\begin{figure*}
\begin{minipage}[t]{0.33\linewidth}
\subfigure{
\centering
\includegraphics[height=1.35in,width=1.7in]{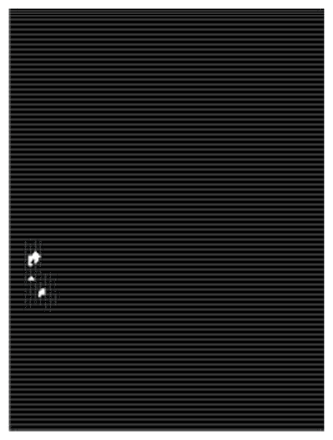}
}
\end{minipage}
\begin{minipage}[t]{0.33\linewidth}
\subfigure{
\centering
\includegraphics[width=1.7in]{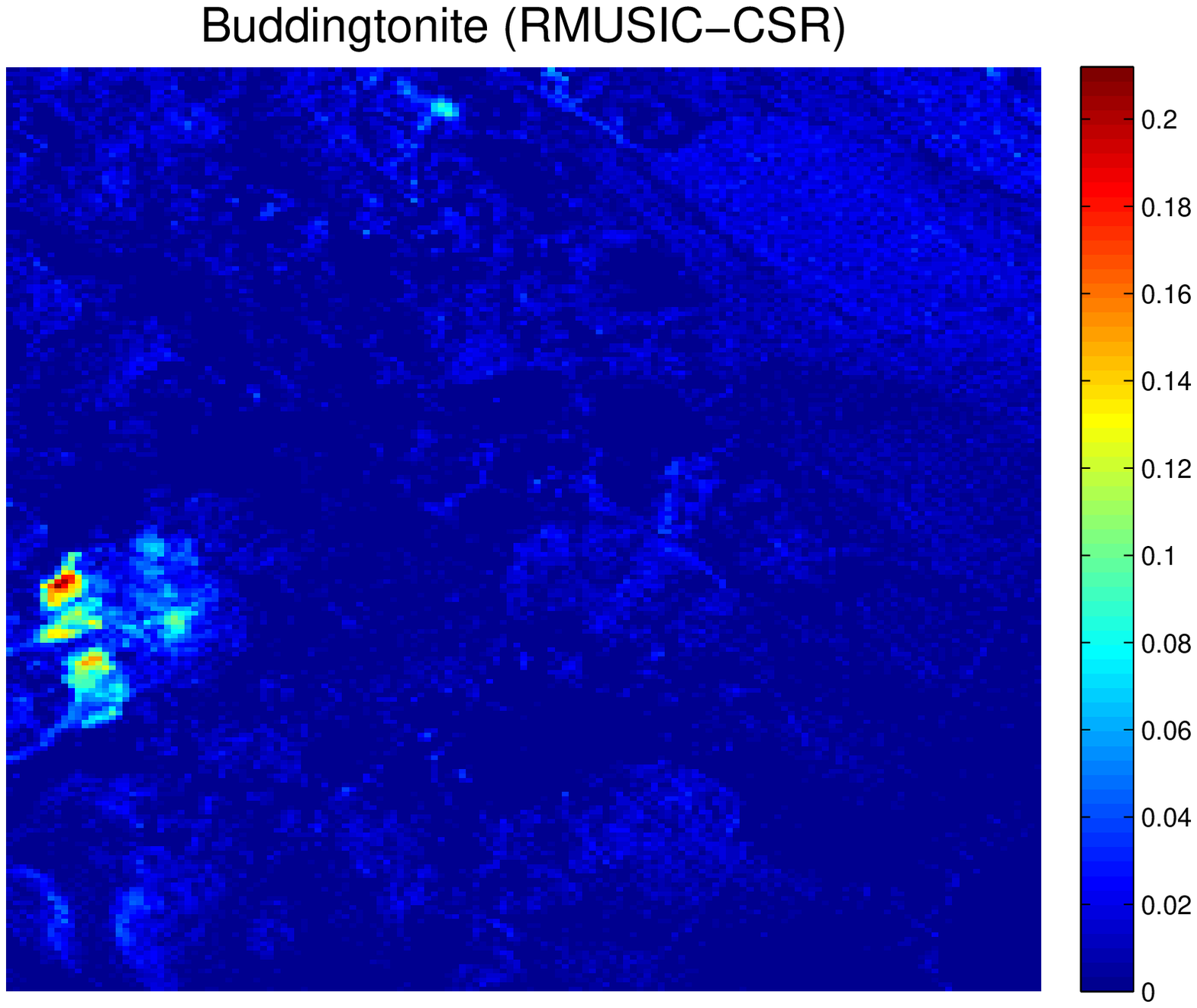}
}
\end{minipage}%
\begin{minipage}[t]{0.33\linewidth}
\subfigure{
\centering
\includegraphics[width=1.7in]{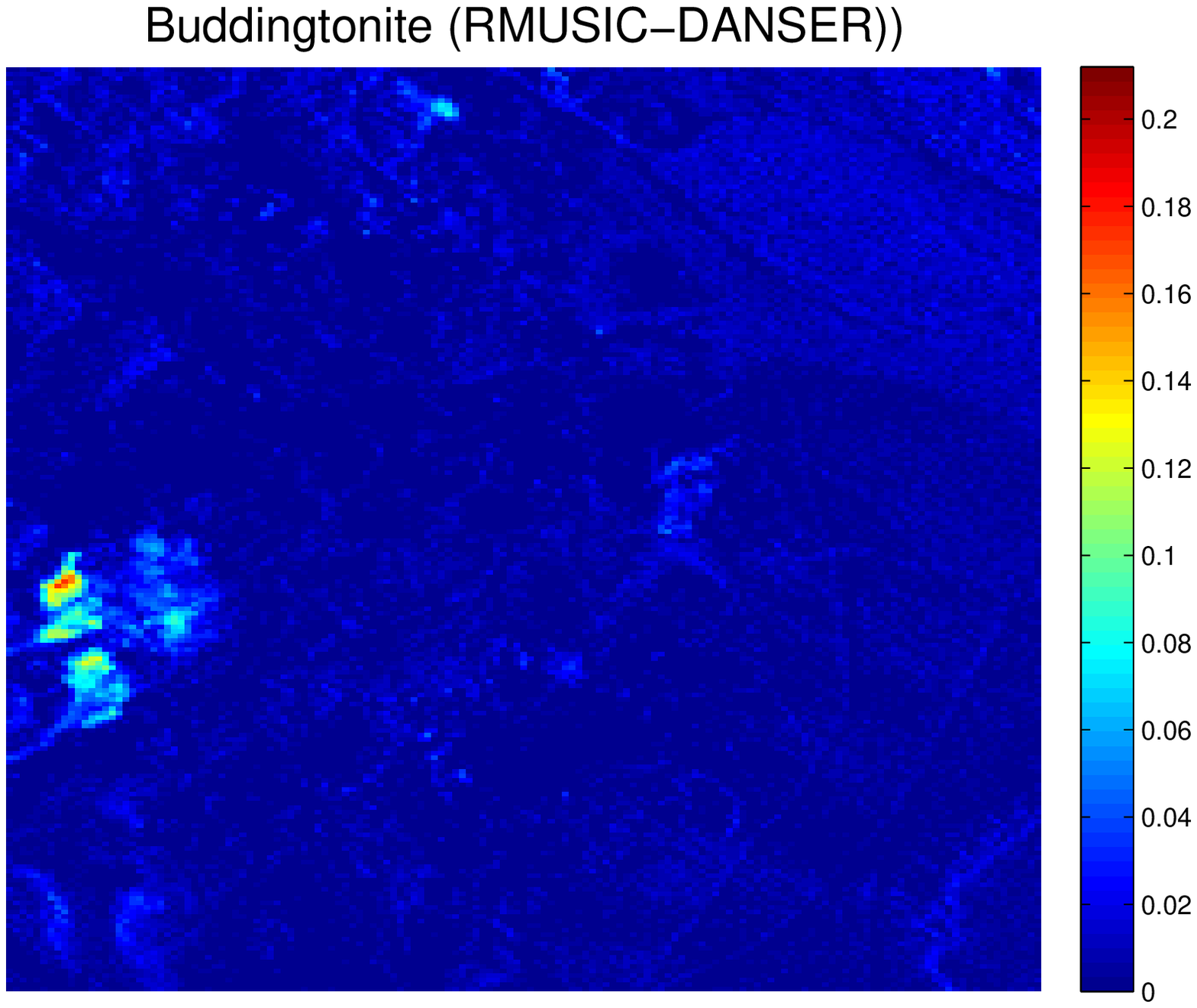}
}
\end{minipage}
\caption{The U.S.G.S. Tetracorder abundance map (left) and the estimated abundance map of the Buddingtonite by RMUSIC-CSR and RMUSIC-DANSER, respectively.}\label{fig:abundance_2}
\end{figure*}

\begin{figure*}
\begin{minipage}[t]{0.33\linewidth}
\subfigure{
\centering
\includegraphics[height=1.35in,width=1.7in]{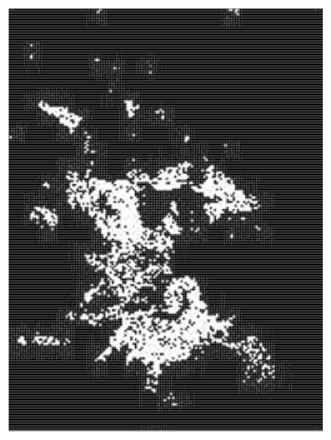}
}
\end{minipage}
\begin{minipage}[t]{0.33\linewidth}
\subfigure{
\centering
\includegraphics[width=1.7in]{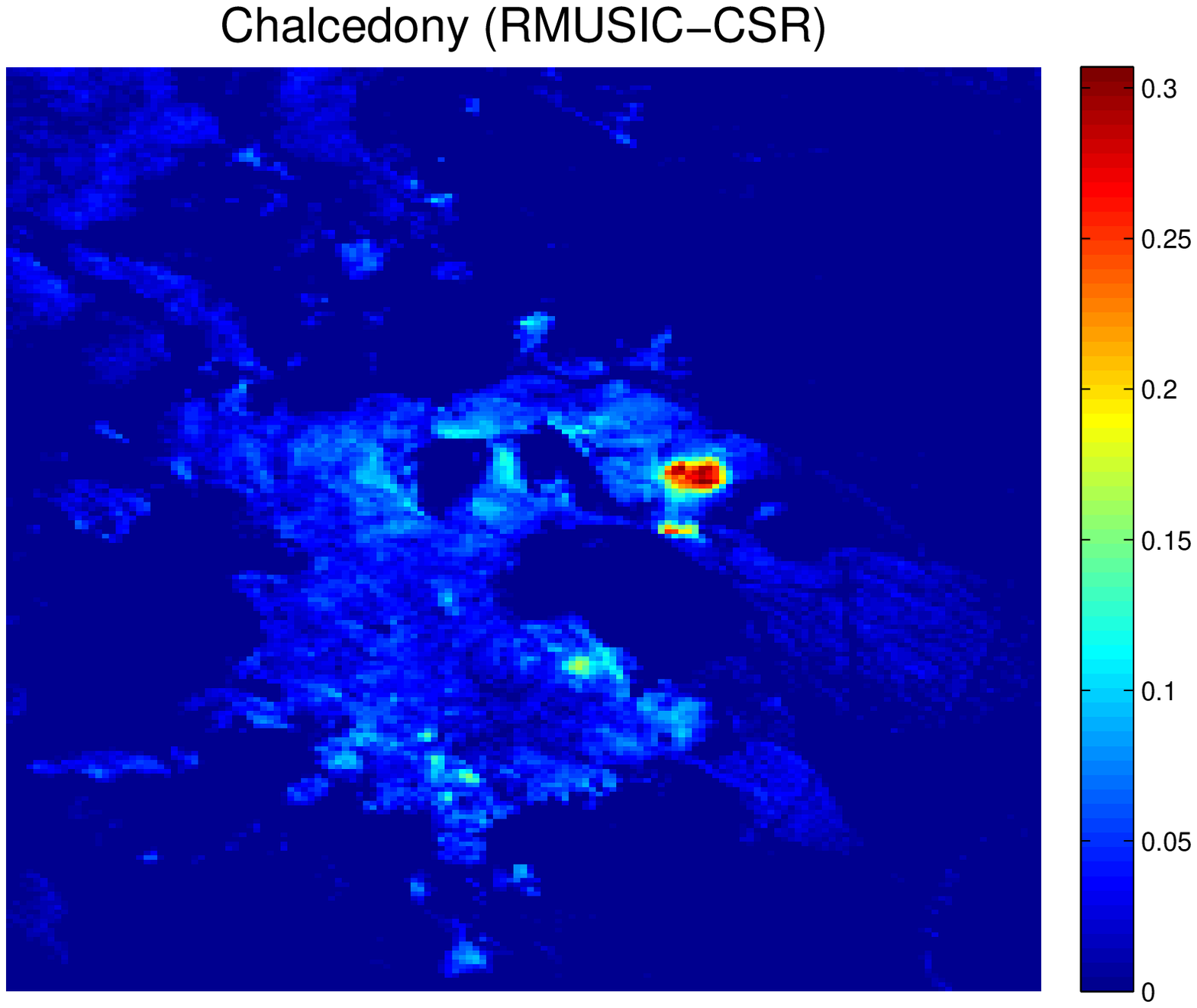}}
\end{minipage}%
\begin{minipage}[t]{0.33\linewidth}
\subfigure{
\centering
\includegraphics[width=1.7in]{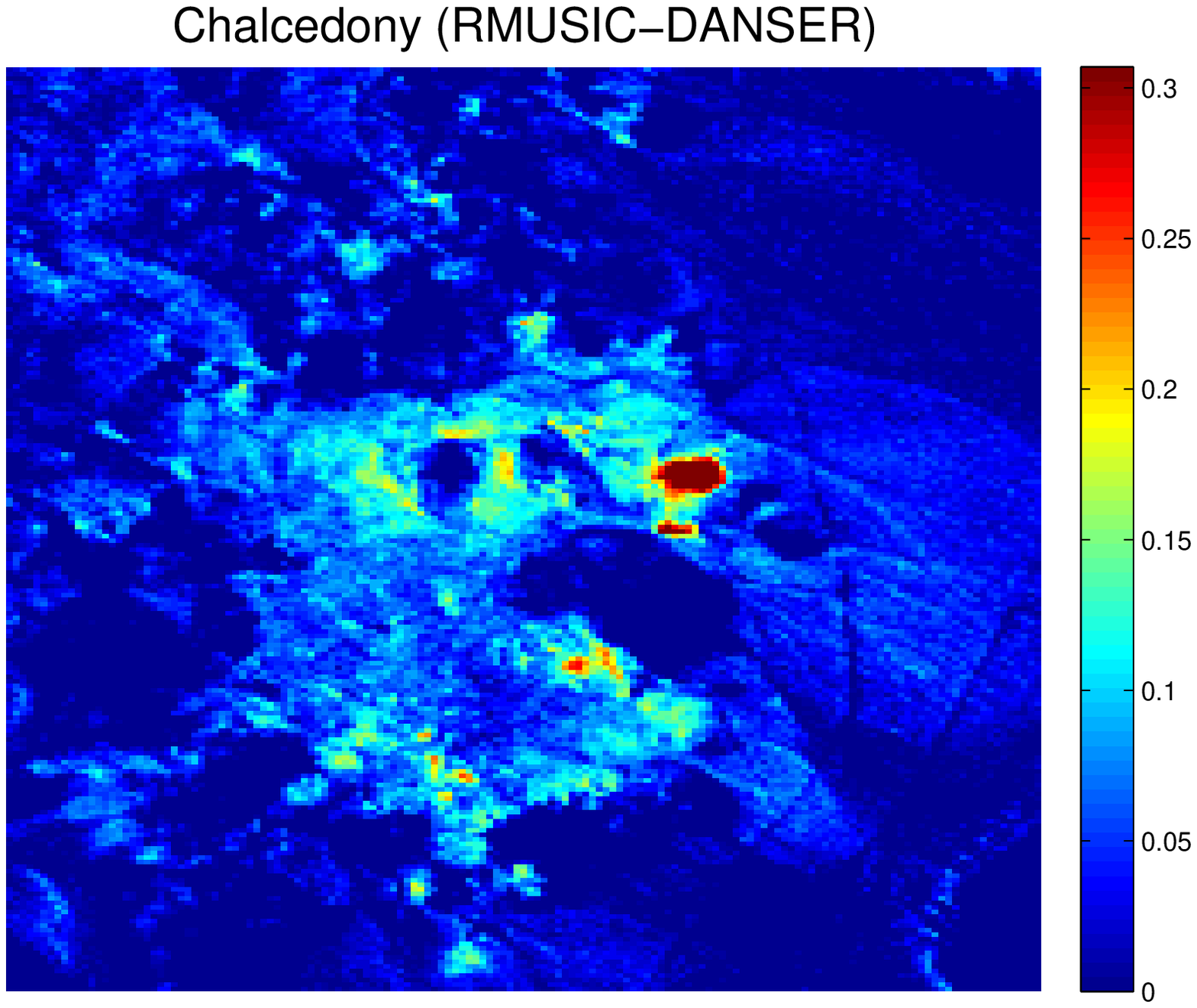}}
\end{minipage}
\caption{The U.S.G.S. Tetracorder abundance map (left) and  estimated abundance map of the Chalcedony by RMUSIC-CSR and RMUSIC-DANSER, respectively.}\label{fig:abundance_3}
\end{figure*}

\begin{figure*}
\begin{minipage}[t]{0.33\linewidth}
\subfigure{
\centering
\includegraphics[height=1.35in,width=1.7in]{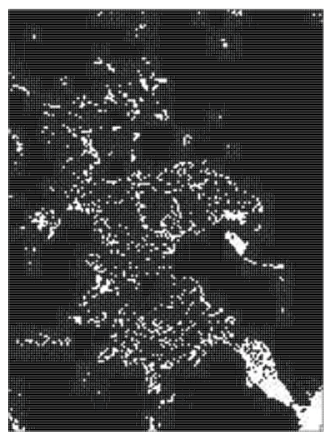}
}
\end{minipage}
\begin{minipage}[t]{0.33\linewidth}
\subfigure{
\centering
\includegraphics[width=1.7in]{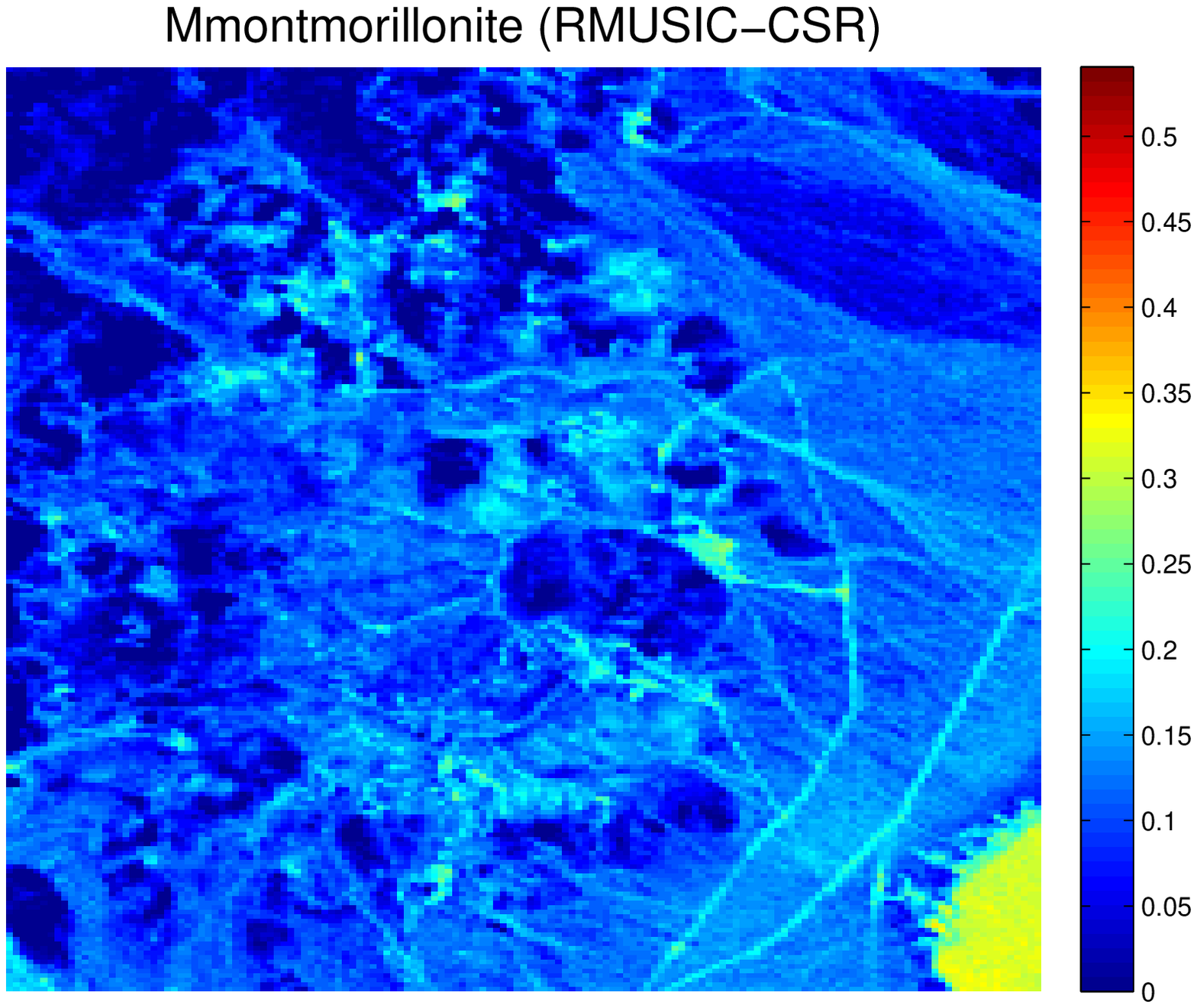}
}
\end{minipage}%
\begin{minipage}[t]{0.33\linewidth}
\subfigure{
\centering
\includegraphics[width=1.7in]{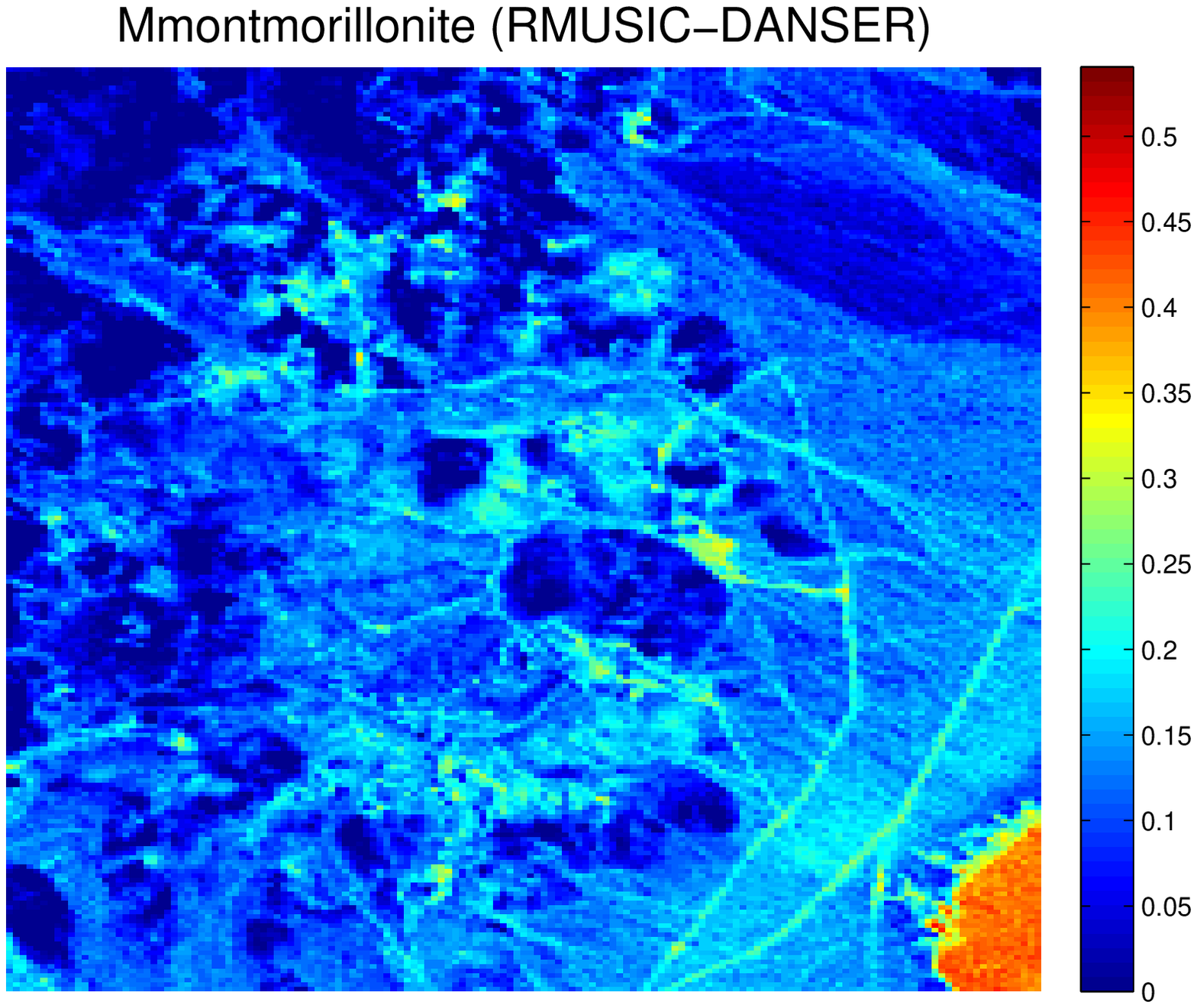}
}
\end{minipage}
\caption{The U.S.G.S. Tetracorder abundance map (left) and  estimated abundance map of the Mmontmorillonite by RMUSIC-CSR and RMUSIC-DANSER, respectively.}\label{fig:abundance_4}
\end{figure*}

We follow the method as in \cite{Iordache2011,Iordache2012TV,Iordache2012collaborative}
to compare the abundance map estimation results of RMUSIC-CSR and RMUSIC-DANSER.
Specifically, we plot the classification maps yielded by the U.S.G.S. Tetracorder software
\cite{clark2003imaging} and the estimated abundance maps of Alunite, Buddingtonite, Chalcedony, and Mmontmorillonite by RMUSIC-CSR and RMUSIC-DANSER in Fig.~\ref{fig:abundance_1} - Fig.~\ref{fig:abundance_4}.
As mentioned in \cite{Iordache2011,Iordache2012TV,Iordache2012collaborative},
the classification maps are based on the older version Cuprite data captured in 1995, while the hyperspectral image was captured in 1997,
which means that the details of the new data may not be fully revealed by the classification maps - but it still makes a good reference for visual evaluation.
We see that for Alunite, Buddingtonite, and Chalcedony,
RMUSIC-CSR and RMUSIC-DANSER yield similar abundance maps.
However, for Chalcedony and Mmontmorillonite, the abundances given by RMUSIC-DANSER generally have stronger intensities all over the area of interest.
Also, by enumerating the nonzero rows of the solution, it is noticed that DANSER identifies $15$ active spectra from the pruned dictionary that consists of 79 spectra, indicating that the number of materials in the subimage is 15. This result is close to that yielded by HySiMe \cite{Hysime}; HySiMe is reliable in estimating the number of materials, and its estimate of this scene is 16.
At the same time, CSR selects 19 active spectra.
This observation also verifies our claim that using $\ell_p$ quasi-norm yields sparser solution.

\section{Conclusion}
In this work, we have developed a dictionary-aided semiblind HU method that takes into account the spectral signature mismatch problem.
We proposed a dictionary-adjusted CSR formulation with a nonconvex collaborative sparsity promoting regularizer.
By a careful reformulation, an alternating optimization algorithm with simple per-iteration updates was proposed.
A new dictionary pruning algorithm based on a spectral mismatch-robust MUSIC criterion was also proposed.
Simulations and real-data experiments showed that the proposed algorithms are promising in improving the HU performance compared to the prior works.

\appendix

\ifplainver
    \section*{Appendix}
    \renewcommand{\thesubsection}{\Alph{subsection}}
\else
    \section{Appendix}
\fi

\subsection{Proof of Proposition~\ref{prop:convergence}}\label{app:convergence}

First, we claim that any limit point of the solution sequence generated by the DANSER algorithm in Algorithm~\ref{Algo:newDANSER} is a stationary point of Problem~\eqref{eq:STLS2} (but not Problem~\eqref{eq:DANSR} at this moment).
The claim is obtained by applying a general alternating optimization (AO) result in \cite[Proposition~2.7.1]{bertsekas1999nonlinear},
which says that every limit point of a solution sequence generated by an AO algorithm is a stationary point of its tackled problem if
each partial optimization problem in AO is strictly convex and has a continuously differentiable objective function within the interior of its feasible set.
In our case, one can see that the partial optimizations of Problem~\eqref{eq:STLS2} w.r.t. ${\bm H}$, ${\bm D}'$, ${\bm c}^1, \ldots, {\bm c}^K$ and $\{{w_k}\}$ satisfy the above condition.

Second, we claim that a stationary point of Problem~\eqref{eq:STLS2} is also a stationary point of Problem~\eqref{eq:DANSR}.
The proof
is as follows.
For notational convenience, let ${\bm X}= [ {\bm C}, {\bm H}^T, ({\bm D}')^T ]$, ${\bm w}=[ w_1,\ldots,w_K ]^T$,
and denote
\begin{align*}
g({\bm X},{\bm w}) & =
h({\bm X}) + \lambda\sum_{k=1}^K \left(w_k\left\|{\bm c}^k\right\|_2^2+\phi_p(w_k)\right) \\
f({\bm X}) & = h({\bm X}) + \lambda \sum_{k=1}^K\left(\|{\bm c}^k\|_{2}^2+\tau\right)^{p/2} 
\end{align*}
as the objective functions of Problem~\eqref{eq:STLS2} and Problem~\eqref{eq:DANSR}, resp., where
\begin{align*}
h({\bm X}) & = \frac{1}{2}\|{\bm Y}-{\bm H}{\bm C}\|_F^2 +\frac{\mu}{2}\|{\bm H}-{\bm D}'\|_F^2.
\end{align*}
Also, recall from the development in Section III.B that
\begin{align*}
f({\bm X}) & = \min_{ {\bm w} \geq {\bm 0} } g({\bm X},{\bm w}).
\end{align*}
Now, let $({\bm X}^\star, {\bm w}^\star)$ be a stationary point of Problem~\eqref{eq:STLS2},
which, by definition, satisfies
\begin{subequations}
\begin{align}
& ( \nabla_{ \bm w } g({\bm X}^\star, {\bm w}^\star) )^T ({\bm w} -{\bm w}^\star)  \leq 0, ~ \forall {\bm w} \geq {\bm 0},  \label{eq:stat_pt_w}\\
&{\rm Tr}\left( ( \nabla_{\bm X} g({\bm X}^\star, {\bm w}^\star))^T ({\bm X}-{\bm X}^\star) \right)\leq 0, ~\forall {\bm X} \in \mathcal{X},  \label{eq:stat_pt_X}
\end{align}
\end{subequations}
where $\nabla_{\bm X} g({\bm X}, {\bm w})$ and $\nabla_{\bm w} g({\bm X}, {\bm w})$ denote the gradient of $g({\bm X}, {\bm w})$ w.r.t. ${\bm X}$ and ${\bm w}$, resp.,
and $\mathcal{X}$ denotes the feasible set of ${\bm X}$ in Problem~\eqref{eq:STLS2} or Problem~\eqref{eq:DANSR}.
From \eqref{eq:stat_pt_w}, we observe that
\begin{align} \label{eq:g_temp}
g({\bm X}^\star,{\bm w}^\star) & = \min_{ {\bm w} \geq {\bm 0} } g({\bm X}^\star,{\bm w}),
\end{align}
and the argument is as follows:
$g$ is strictly convex w.r.t. ${\bm w} \geq {\bm 0}$ by Lemma~\ref{lem:conjugate}; and
as a result of the optimality conditions of convex optimization, \eqref{eq:stat_pt_w} holds if and only if ${\bm w}^\star$ is the optimal solution to $\min_{\bm w \geq \bm 0} g({\bm X}^\star,{\bm w})$.
Eq.~\eqref{eq:g_temp} implies that $f({\bm X}^\star)  = g({\bm X}^\star,{\bm w}^\star)$.
Consequently, we can rewrite \eqref{eq:stat_pt_X} as
\begin{align*}
&{\rm Tr}\left( ( \nabla_{\bm X} f({\bm X}^\star))^T ({\bm X}-{\bm X}^\star) \right)\leq 0, ~\forall {\bm X} \in \mathcal{X}.
\end{align*}
The above equation is identical to the definition for ${\bm X}^\star$ to be a stationary point of Problem~\eqref{eq:DANSR}.
Hence, we have proven that for any stationary point $({\bm X}^\star, {\bm w}^\star)$ of Problem~\eqref{eq:STLS2}, the part ${\bm X}^\star$ is a stationary point of Problem~\eqref{eq:DANSR}.

Finally, combining the above two claims leads to the conclusion in Proposition~\ref{prop:convergence}.

\subsection{Proof of Proposition~\ref{prop:etastar}}\label{app:rmusic}

Recall that we aim at solving
\begin{equation}\label{eq:one}
\min_{\|{\bm \xi}\|_2\leq \epsilon}~\eta_k({\bm \xi}),
\end{equation}
where
\begin{equation}
\eta_k({\bm \xi}) = \frac{\left\|{\bm P}^\perp_{{\bm U}_S}({\bm d}_k-{\bm \xi})\right\|_2}{
 \| {\bm P}_{{\bm U}_S}({\bm d}_k-{\bm \xi})
 \|_2 }.
\end{equation}
By the triangle inequality, we have
\begin{equation}
\eta_k({\bm \xi}) \geq \frac{\left|\left\|{\bm P}^\perp_{{\bm U}_S}{\bm d}_k\right\|_2-\left\|{\bm P}^\perp_{{\bm U}_S}{\bm \xi}\right\|_2\right|}{\left\|{\bm P}_{{\bm U}_S}{\bm d}_k\right\|_2+\left\|{\bm P}_{{\bm U}_S}{\bm \xi}\right\|_2},
\end{equation}
where equality holds if and only if i) ${\bm P}^{\perp}_{{\bm U}_S}{\bm \xi}=\beta {\bm P}^{\perp}_{{\bm U}_S}{\bm d}_k$, $\beta\geq 0$, and ii)
${\bm P}_{{\bm U}_S}{\bm \xi}=\alpha {\bm P}_{{\bm U}_S}{\bm d}_k$, $\alpha\geq 0$.
The two conditions above can be satisfied simultaneously by setting
\begin{equation}\label{eq:four}
{\bm \xi} = -\frac{\alpha}{\|{\bm P}_{{\bm U}_S}{\bm d}_k\|_2}{\bm P}_{{\bm U}_S}{\bm d}_k+\frac{\beta}{\|{\bm P}^{\perp}_{{\bm U}_S}{\bm d}_k\|_2}{\bm P}^{\perp}_{{\bm U}_S}{\bm d}_k
\end{equation}
for some $\alpha,\beta\geq 0$.
Also, note that $\|{\bm \xi}\|_2\leq \epsilon$ is equivalent to
\begin{equation}\label{eq:five}
\alpha^2 + \beta^2\leq \epsilon^2.
\end{equation}
Substituting \eqref{eq:four} into $\eta_k({\bm \xi})$, and by noting \eqref{eq:five},
we recast Problem~\eqref{eq:one} as
\begin{equation}\label{eq:six}
\min_{\begin{subarray}{c}\alpha,\beta\geq 0\\ \alpha^2 + \beta^2\leq \epsilon^2\end{subarray}}\quad\frac{\left|\left\|{\bm P}^\perp_{{\bm U}_S}{\bm d}_k\right\|_2-\beta\right|}{\left\|{\bm P}_{{\bm U}_S}{\bm d}_k\right\|_2+\alpha}.
\end{equation}
Consider two cases, namely, i) $\left\|{\bm P}^\perp_{{\bm U}_S}{\bm d}_k\right\|_2\leq \epsilon^2$, and ii) $\left\|{\bm P}^\perp_{{\bm U}_S}{\bm d}_k\right\|_2> \epsilon^2$.
For case i), the optimal $\beta$ is $\beta=\left\|{\bm P}^\perp_{{\bm U}_S}{\bm d}_k\right\|_2$, and the optimal $\alpha$ may take any value in
$\left[0,\sqrt{\epsilon^2-\|{\bm P}^\perp_{{\bm U}_S}{\bm d}_k\|_2^2}\right]$.
For case ii), we observe the following: fixing $\beta$, $\alpha$ should be made as large as possible so as to reduce the objective value.
Hence, we can substitute $\alpha = \sqrt{\epsilon^2-\beta^2}$ (the largest possible $\alpha$ fixing $\beta$) into Problem~\eqref{eq:six} and simplify the problem to
\begin{equation}
         \eta_k^\star=\min_{0\leq \beta \leq \epsilon}~\frac{\left| \|{\bm P}^{\perp}_{{\bm U}_S}{\bm d}_k\|_2 - \beta \right|}{\|{\bm P}_{{\bm U}_S}{\bm d}_k\|_2 + \sqrt{\epsilon^2-\beta^2}},
\end{equation}
which is exactly Problem~\eqref{eq:etastar}.

\bibliography{refs_music}

\begin{thebibliography}{10}
\providecommand{\url}[1]{#1}
\csname url@samestyle\endcsname
\providecommand{\newblock}{\relax}
\providecommand{\bibinfo}[2]{#2}
\providecommand{\BIBentrySTDinterwordspacing}{\spaceskip=0pt\relax}
\providecommand{\BIBentryALTinterwordstretchfactor}{4}
\providecommand{\BIBentryALTinterwordspacing}{\spaceskip=\fontdimen2\font plus
\BIBentryALTinterwordstretchfactor\fontdimen3\font minus
  \fontdimen4\font\relax}
\providecommand{\BIBforeignlanguage}[2]{{%
\expandafter\ifx\csname l@#1\endcsname\relax
\typeout{** WARNING: IEEEtran.bst: No hyphenation pattern has been}%
\typeout{** loaded for the language `#1'. Using the pattern for}%
\typeout{** the default language instead.}%
\else
\language=\csname l@#1\endcsname
\fi
#2}}
\providecommand{\BIBdecl}{\relax}
\BIBdecl

\bibitem{fu2013greedy}
X.~Fu, W.-K. Ma, T.-H. Chan, J.~M. Bioucas-Dias, and M.-D. Iordache, ``Greedy
  algorithms for pure pixels identification in hyperspectral unmixing: A
  multiple-measurement vector viewpoint,'' \emph{{\rm in} Proc. EUSIPCO 2013},
  2013.

\bibitem{Bioucas2012}
J.~Bioucas-Dias, A.~Plaza, N.~Dobigeon, M.~Parente, Q.~Du, P.~Gader, and
  J.~Chanussot, ``Hyperspectral unmixing overview: Geometrical, statistical,
  and sparse regression-based approaches,'' \emph{IEEE J. Sel. Topics Appl.
  Earth Observ.}, vol.~5, no.~2, pp. 354--379, 2012.

\bibitem{Ma2014}
W.-K. Ma, J.~Bioucas-Dias, T.-H. Chan, N.~Gillis, P.~Gader, A.~Plaza,
  A.~Ambikapathi, and C.-Y. Chi, ``A signal processing perspective on
  hyperspectral unmixing,'' \emph{IEEE Signal Process. Mag.}, vol.~31, no.~1,
  pp. 67--81, Jan 2014.

\bibitem{USGS2007}
\BIBentryALTinterwordspacing
R.~Clark, G.~Swayze, R.~Wise, E.~Livo, T.~Hoefen, R.~Kokaly, and S.~Sutley,
  ``{USGS} digital spectral library splib06a: {U.S. Geological Survey, Digital
  Data Series 231},'' 2007. [Online]. Available:
  \url{http://speclab.cr.usgs.gov/spectral.lib06}
\BIBentrySTDinterwordspacing

\bibitem{Iordache2011}
M.-D. Iordache, J.~Bioucas-Dias, and A.~Plaza, ``Sparse unmixing of
  hyperspectral data,'' \emph{IEEE Trans. Geosci. Remote Sens.}, vol.~49,
  no.~6, pp. 2014--2039, 2011.

\bibitem{Iordache2012TV}
------, ``Total variation spatial regularization for sparse hyperspectral
  unmixing,'' \emph{IEEE Trans. Geosci. Remote Sens.}, vol.~50, no.~11, pp.
  4484--4502, 2012.

\bibitem{Iordache2012collaborative}
------, ``Collaborative sparse regression for hyperspectral unmixing,''
  \emph{IEEE Trans. Geosci. Remote Sens.}, vol.~52, no.~1, pp. 341--354, Jan
  2014.

\bibitem{joseMUSIC2013}
------, ``{M}{U}{S}{I}{C}-{C}{S}{R}: Hyperspectral unmixing via multiple signal
  classification and collaborative sparse regression,'' \emph{IEEE Trans.
  Geosci. Remote Sens.}, vol.~52, no.~7, pp. 4364--4382, July 2014.

\bibitem{Tropp2006pt2}
J.~Tropp, ``Algorithms for simultaneous sparse approximation. {Part II}: Convex
  relaxation,'' \emph{Signal Process.}, vol.~86, no.~3, pp. 589--602, 2006.

\bibitem{Eldar2010average}
Y.~Eldar and H.~Rauhut, ``Average case analysis of multichannel sparse recovery
  using convex relaxation,'' \emph{IEEE Trans. Inf. Theory}, vol.~56, no.~1,
  pp. 505--519, 2010.

\bibitem{Chen2006}
J.~Chen and X.~Huo, ``Theoretical results on sparse representations of
  multiple-measurement vectors,'' \emph{IEEE Trans. Signal Process.}, vol.~54,
  no.~12, pp. 4634 --4643, Dec. 2006.

\bibitem{tropp2006just}
J.~A. Tropp, ``Just relax: Convex programming methods for identifying sparse
  signals in noise,'' \emph{IEEE Trans. Inf. Theory}, vol.~52, no.~3, pp.
  1030--1051, 2006.

\bibitem{schmidt1986multiple}
R.~O. Schmidt, ``Multiple emitter location and signal parameter estimation,''
  \emph{IEEE Trans. Antennas Propag.}, vol.~34, no.~3, pp. 276--280, 1986.

\bibitem{kim2010compressive}
J.~M. Kim, O.~K. Lee, and J.~C. Ye, ``Compressive music: Revisiting the link
  between compressive sensing and array signal processing,'' \emph{IEEE Trans.
  Inf. Theory}, vol.~58, no.~1, pp. 278--301, Jan 2012.

\bibitem{somers2011endmember}
B.~Somers, G.~P. Asner, L.~Tits, and P.~Coppin, ``Endmember variability in
  spectral mixture analysis: A review,'' \emph{Remote Sensing of Environment},
  vol. 115, no.~7, pp. 1603--1616, 2011.

\bibitem{Chartrand2008}
R.~Chartrand and W.~Yin, ``Iteratively reweighted algorithms for compressive
  sensing,'' in \emph{Proc. ICASSP 2008.}, 31 2008-April 4 2008, pp. 3869
  --3872.

\bibitem{Rao1999}
B.~Rao and K.~Kreutz-Delgado, ``An affine scaling methodology for best basis
  selection,'' \emph{IEEE Trans. Signal Process.}, vol.~47, no.~1, pp. 187
  --200, jan 1999.

\bibitem{chartrand2008restricted}
R.~Chartrand and V.~Staneva, ``Restricted isometry properties and nonconvex
  compressive sensing,'' \emph{Inverse Problems}, vol.~24, no.~3, p. 035020,
  2008.

\bibitem{zhu2011sparsity}
H.~Zhu, G.~Leus, and G.~Giannakis, ``Sparsity-cognizant total least-squares for
  perturbed compressive sampling,'' \emph{IEEE Trans. Signal Process.},
  vol.~59, no.~5, pp. 2002--2016, May 2011.

\bibitem{gribonval2012blind}
R.~Gribonval, G.~Chardon, and L.~Daudet, ``Blind calibration for compressed
  sensing by convex optimization,'' in \emph{Proc. IEEE ICASSP 2012}, 2012, pp.
  2713--2716.

\bibitem{bilen2013blind}
C.~Bilen, G.~Puy, R.~Gribonval, and L.~Daudet, ``Blind sensor calibration in
  sparse recovery using convex optimization,'' in \emph{SAMPTA-10th
  International Conference on Sampling Theory and Applications-2013}, 2013.

\bibitem{tan2014joint}
Z.~Tan, P.~Yang, and A.~Nehorai, ``Joint sparse recovery method for compressed
  sensing with structured dictionary mismatches,'' \emph{IEEE Trans. Signal
  Process.}, vol.~62, no.~19, pp. 4997--5008, Oct 2014.

\bibitem{chen2013sparse}
F.~Chen and Y.~Zhang, ``Sparse hyperspectral unmixing based on constrained
  $\ell_p$ - $\ell_2$ optimization,'' \emph{IEEE Geosci. Remote Sens. Lett.},
  vol.~10, no.~5, pp. 1142--1146, Sept 2013.

\bibitem{Hysime}
J.~Bioucas-Dias and J.~Nascimento, ``Hyperspectral subspace identification,''
  \emph{IEEE Trans. Geosci. Remote Sens.}, vol.~46, no.~8, pp. 2435--2445,
  2008.

\bibitem{shen2013exact}
Y.~Shen, J.~Fang, and H.~Li, ``Exact reconstruction analysis of log-sum
  minimization for compressed sensing,'' \emph{IEEE Signal Process. Lett.},
  vol.~20, no.~12, pp. 1223--1226, 2013.

\bibitem{courant1943variational}
R.~Courant \emph{et~al.}, ``Variational methods for the solution of problems of
  equilibrium and vibrations,'' \emph{Bull. Amer. Math. Soc}, vol.~49, no.~1,
  pp. 1--23, 1943.

\bibitem{wang2008new}
Y.~Wang, J.~Yang, W.~Yin, and Y.~Zhang, ``A new alternating minimization
  algorithm for total variation image reconstruction,'' \emph{SIAM Journal on
  Imaging Sciences}, vol.~1, no.~3, pp. 248--272, 2008.

\bibitem{xiao2010fast}
\BIBentryALTinterwordspacing
Y.~Xiao and J.~Yang, ``{A Fast Algorithm for Total Variation Image
  Reconstruction from Random Projections},'' Tech. Rep. arXiv:1001.1774, Jan
  2010. [Online]. Available: \url{http://cds.cern.ch/record/1232976}
\BIBentrySTDinterwordspacing

\bibitem{fu2015joint}
X.~Fu, K.~Huang, W.-K. Ma, N.~D. Sidiropoulos, and B.~Rasmus, ``Joint slab
  selection and low-rank tensor decomposition with applications,'' \emph{{\rm
  submitted to} IEEE Trans. Signal Process.}, 2015.

\bibitem{geman1992constrained}
D.~Geman and G.~Reynolds, ``Constrained restoration and the recovery of
  discontinuities,'' \emph{IEEE Trans. Pattern Anal. Mach. Intell.}, vol.~14,
  no.~3, pp. 367--383, 1992.

\bibitem{idier2001convex}
J.~Idier, ``Convex half-quadratic criteria and interacting auxiliary variables
  for image restoration,'' \emph{IEEE Trans. Image Process.}, vol.~10, no.~7,
  pp. 1001--1009, 2001.

\bibitem{cichocki2009fast}
A.~Cichocki and P.~Anh-Huy, ``Fast local algorithms for large scale nonnegative
  matrix and tensor factorizations,'' \emph{IEICE Transactions on Fundamentals
  of Electronics, Communications and Computer Sciences}, vol.~92, no.~3, pp.
  708--721, 2009.

\bibitem{chemometrics1998least}
R.~Bro and N.~D. Sidiropoulos, ``Least squares algorithms under unimodality and
  i non-negativity constraints,'' \emph{J. Chemometrics}, vol.~12, pp.
  223--247, 1998.

\bibitem{zhang2008quadratic}
A.-L. Zhang, ``Quadratic fractional programming problems with quadratic
  constraints,'' Ph.D. dissertation, Kyoto University, 2008.

\bibitem{dinkelbach1967nonlinear}
W.~Dinkelbach, ``On nonlinear fractional programming,'' \emph{Management
  Science}, vol.~13, no.~7, pp. 492--498, 1967.

\bibitem{clark2003imaging}
R.~N. Clark, G.~A. Swayze, K.~E. Livo, R.~F. Kokaly, S.~J. Sutley, J.~B.
  Dalton, R.~R. McDougal, and C.~A. Gent, ``Imaging spectroscopy: Earth and
  planetary remote sensing with the usgs tetracorder and expert systems,''
  \emph{Journal of Geophysical Research: Planets (1991--2012)}, vol. 108, no.
  E12, 2003.

\bibitem{bertsekas1999nonlinear}
D.~Bertsekas, \emph{Nonlinear programming}.\hskip 1em plus 0.5em minus
  0.4em\relax Athena Scientific, 1999.

\end{thebibliography}

\end{document}